\documentclass[journal]{IEEEtran}
\usepackage{cite}

\ifCLASSINFOpdf
  \usepackage[pdftex]{graphicx}
\else
  \usepackage[dvips]{graphicx}
\fi
\usepackage{amsmath}

\DeclareMathOperator*{\argmin}{arg\,min}
\usepackage{amsfonts}
\usepackage[linesnumbered,ruled]{algorithm2e}

\usepackage{array}

\usepackage[caption=false,font=normalsize,labelfont=sf,textfont=sf]{subfig}
\usepackage{url}

\usepackage{microtype}
\usepackage{multirow}

\begin{document}
%
\title{Improved Regularity Model-based EDA for Many-objective Optimization}
%
%
%

\author{Yanan~Sun,~\IEEEmembership{Member},~\IEEEmembership{IEEE},
        Gary~G.~Yen,~\IEEEmembership{Fellow},~\IEEEmembership{IEEE},
        and~Zhang~Yi,~\IEEEmembership{Fellow},~\IEEEmembership{IEEE}
        \thanks{This work is supported in part by the China Scholarship Council under Grant 201506240048;  in part by the Miaozi Project in Science and Technology Innovation Program of Sichuan Province under Grant 16-YCG061, China; in part by the National Natural Science Foundation of China for Distinguished Young Scholar under Grant 61622504; and in part by the National
        	Natural Science Foundation of China under Grant 61432012 and Grant U1435213.~\emph{(Corresponding author:
Gary G. Yen.)}}
\thanks{Yanan Sun is with the College of Computer Science, Sichuan University, Chengdu 610065 CHINA and with the School of Engineering and Computer Science, Victoria University of Wellington, Wellington 6140 NEW ZEALAND (e-mail:yanan.sun@ecs.vuw.ac.nz).}
\thanks{Gary G. Yen is with the School of Electrical and Computer Engineering, Oklahoma State University, Stillwater, OK 74078 USA (e-mail:gyen@okstate.edu).}
\thanks{Zhang Yi is with the College of Computer Science, Sichuan University, Chengdu 610065 CHINA (e-mail:zhangyi@scu.edu.cn).}
}

\maketitle

\begin{abstract}
The performance of multi-objective evolutionary algorithms deteriorates appreciably in solving many-objective optimization problems which encompass more than three objectives. One of the known rationales is the loss of selection pressure which leads to the selected parents not generating promising offspring towards Pareto-optimal front with diversity. Estimation of distribution algorithms sample new solutions with a probabilistic model built from the statistics extracting over the existing solutions so as to mitigate the adverse impact of genetic operators. In this paper, an improved regularity-based estimation of distribution algorithm is proposed to effectively tackle unconstrained many-objective optimization problems. In the proposed algorithm, \emph{diversity repairing mechanism} is utilized to mend the areas where need non-dominated solutions with a closer proximity to the Pareto-optimal front. Then \emph{favorable solutions} are generated by the model built from the regularity of the solutions surrounding a group of representatives. These two steps collectively enhance the selection pressure which gives rise to the superior convergence of the proposed algorithm. In addition, dimension reduction technique is employed in the decision space to speed up the estimation search of the proposed algorithm. Finally, by assigning the Pareto-optimal solutions to the uniformly distributed reference vectors, a set of solutions with excellent diversity and convergence is obtained. To measure the performance, NSGA-III, GrEA, MOEA/D, HypE, MBN-EDA, and RM-MEDA are selected to perform comparison experiments over DTLZ and DTLZ$^-$ test suites with $3$-, $5$-, $8$-, $10$-, and $15$-objective. Experimental results quantified by the selected performance metrics reveal that the proposed algorithm shows considerable competitiveness in addressing unconstrained many-objective optimization problems.
\end{abstract}

\begin{IEEEkeywords}
Estimation distribution algorithm~(EDA), many-objective evolutionary algorithm~(MaOEA), regularity-based EDA, diversity repairing, decision space dimension reduction.
\end{IEEEkeywords}

%
\IEEEpeerreviewmaketitle

\section{Introduction}
\label{section_1}
\IEEEPARstart{M}{any-objective} optimization problems (MaOPs) concern solving $M$ conflicting objectives simultaneously where $M$ is greater than three~\cite{farina2002optimal}. Generally, an MaOP has the following formulation described by Equation~(\ref{equation_mops})
\label{section_introduction}
\begin{equation}
\label{equation_mops}
\setlength{\arraycolsep}{1pt}
\renewcommand{\arraystretch}{1.5}
\left\{
\begin{array}{c}
  f(x)=\left[f_1(x),\cdots,f_M(x)\right] \\
  s.t.~~~x\in \Lambda
\end{array}
\right.
\end{equation}
where $\Lambda\in \mathbb{R}^n$ is the decision space, $f:\Lambda \rightarrow \Omega \in \mathbb{R}^M$ is the objective space. Without loss of generality, it is assumed that $f(x)$ is a minimization problem in which $f_1(x),\cdots, f_M(x)$ are to be minimized. Because of MaOPs widely existing in many real-world applications, such as management in land exploitation with $14$-objective~\cite{chikumbo2012approximating}, calibration problems of automotive engine with $10$-objective~\cite{lygoe2013real}, to name a few, there is a strong incentive for efficiently and effectively solving MaOPs.

In MaOPs, there is no single perfect solution that optimizes all of the objectives at the same time but a set of Pareto-optimal solutions in which each individual is non-dominated with respect to each other. In addition, all the Pareto-optimal solutions constitute the Pareto-optimal set~(PS) in the decision space while the image of PS produces a Pareto-optimal front~(PF) in the objective space. Commonly, the goal in solving MaOPs is to obtain a limit number of Pareto-optimal solutions, which are uniformly distributed in PF, where a decision-maker can delegate a solution based on his or her preference. Among all the approaches for handing MaOPs, evolutionary algorithms are considered preferable because of the searching power exerted in these population-based meta-heuristics. During the past several decades, numerous multi-objective evolutionary algorithms (MOEAs), such as elitist non-dominated sorting genetic algorithm (NSGA-II)~\cite{deb2002fast}, advanced version of strength Pareto evolutionary algorithm (SPEA2)~\cite{zitzler2001spea2}, have been developed for dealing with multi-objective optimization problems (MOPs) in which at most three objectives are to be optimized simultaneously. However, their performance degraded drastically in addressing MaOPs~\cite{knowles2007quantifying}. The main reason is the loss of selection pressure which is caused by the dominance resistance~(DR)~\cite{fonseca1998multiobjective} and the curse of dimensionality~\cite{purshouse2007evolutionary} phenomena. To be specific, DR refers to a large proportion of solutions in which individuals are the best in one or very few objectives but far worse in others, and these solutions cannot be discriminated by the original Pareto domination principle. Then density-based secondary measurement is activated to decide which solutions are allowed to survive in the next generation~\cite{li2015many}. Because of the behavior influenced by DR pointed out in~\cite{wagner2007pareto}, the selected solutions do not necessarily converge to the PF~\cite{purshouse2007evolutionary}. To this end, various many-objective evolutionary algorithms (MaOEAs) for tackling MaOPs have been developed\footnote{The algorithms which was designed originally for MOPs while is extended for MaOPs are also categorized to MaOEAs in this paper.}, such as multi-objective evolutionary algorithm based on decomposition (MOEA/D)~\cite{zhang2007moea}, hypervolume-based many-objective optimization algorithm (HypE)~\cite{bader2011hype}, grid-based evolutionary algorithm for many-objective optimization (GrEA)~\cite{yang2013grid}, many-objective optimization algorithm using reference point based non-dominated sorting (NSGA-III)~\cite{deb2014evolutionary}, and etc.\cite{li2015many,trivedi2017survey,sun2017global,sun2017reference}\footnote{Typically, these MaOEAs can be classified into three basic categories: Dominance-based, Decomposition-based and Hypervolume-based. MOEA/D and HypE are from the second and third categories, respectively. NSGA-III and GrEA are the hybridization of the first and the second categories. }. More precisely, MOEA/D employs the decomposition-based approach to construct a set of single objective problems by aggregating objectives considered in the original MaOP with different predefined weight vectors. New solutions are generated within a sub-region and diversity is maintained by the uniformly distributed weight vectors. The promising solutions are selected in HypE based on their fitness which is assigned by the corresponding contribution in hypervolume measure. As the computation of exact hypervolume is prohibitive, Monte Carlo simulation is employed to address this limitation. GrEA utilizes grid-based approach to show better performance in solving MaOPs by introducing the grid-based fitness comparison to relax the Pareto-based dominance relationship and grid-metrics to improve the diversity.  Compared to NSGA-II, the improvement of NSGA-III is in the diversity mechanism by assigning the solutions to a set of uniformly distributed reference vectors. In summary, these state-of-the-art MaOEAs mentioned above mainly contemplate on two distinct issues: 1) reform the comparison manner in the traditional dominance relationship, such as HypE and GrEA and 2) apply new designs to reinforce the diversity, such as MOEA/D and NSGA-III.
\begin{algorithm}
  \caption{Framework of An EDA}
  \label{alg_framework_eda}
   $t\leftarrow 0$\;
   $P_t\leftarrow$ Randomly initialize the population\;
   \While{termination is not satisfied}
   {
    $M\leftarrow$ Built probabilistic models from $P_t$\;
    $t\leftarrow t+1$\;
    $U_t\leftarrow$ Generate offspring from $M$\;
    $P_t\leftarrow$ Select promising solutions from $U_t\cup P_{t-1}$\;
   }
   \textbf{Return} $P_t$.
\end{algorithm}

It is highly expected that individuals generated by selected parents with the crossover and mutation operators would march towards the PF in a MOP. However, this will not be the case in MaOPs due to the DR phenomenon. Specifically, if the parents are neighbors of DR solutions their offspring are also DR solutions. Otherwise, the newly generated solutions are not necessarily better than their parents who stand at a large space of many-objectives with a remote distance. This can be seen as the inefficiency of existing genetic operators for MaOPs~\cite{purshouse2007evolutionary}. Moreover, Deb \emph{et al.} in~\cite{deb2006multi} concluded that the performances of MOEAs are significantly influenced by the genetic operators which cannot ensure to generate promising offspring. Furthermore, the parameters in genetic operators need empirically configured. For example, the distribution index of SBX in NSGA-III needs to be set at a larger number. For this purpose, researchers have developed estimation of distribution algorithms (EDAs) to tackle optimization problems~\cite{pelikan2006multiobjective,zhang2008rm,karshenas2014multiobjective} by generating new solutions without involving the traditional genetic operators, but probabilistic models which are built based on the statistics of the visited solutions. A general framework of EDAs is illustrated in Algorithm~\ref{alg_framework_eda}. Typically, the EDAs-based MOEAs are broadly classified into two categories based on their estimation models.

The first category covers the Bayesian network-based EDAs. For example, multi-objective Bayesian optimization algorithm~\cite{khan2002multi} utilized the Bayesian optimization algorithm (BOA) to build a Bayesian network as its model for generating offspring. A related work was investigated in~\cite{schwarz2001multiobjective} to predict the model by strengthening Pareto ranking approach~\cite{zitzler1999multiobjective} and BOA. Furthermore, Laumanns \emph{et al.} in~\cite{laumanns2002bayesian} proposed a Bayesian multi-objective optimization algorithm whose model is built over the solutions selected by $\epsilon$-Pareto ranking method~\cite{laumanns2002combining}. In addition, an improved non-dominated sorting approach was employed by decision tree-based multi-objective EDA~\cite{zhong2007decision} to select a subset of solutions serving for a regression decision tree to learn the model. Recently, the multi-dimensional Bayesian network (MBN-EDA) was proposed in~\cite{karshenas2014multiobjective} specifically for addressing MaOPs.

The other category is often known as the mixture probability model-based EDAs. Examples include the multi-objective mixture-based iterated density estimation evolutionary algorithm~\cite{bosman2002multi} employing the mixed probability distributions to sample well-distributed solutions; and the multi-objective Parzen-based EDA~\cite{costa2003moped} learning from the Gaussian and Cauchy kernels to build its models. In~\cite{pelikan2005multiobjective}, the multi-objective hierarchical Bayesian optimization algorithm was designed by the mixture Bayesian network-based probabilistic model for discrete multi-objective optimization problems. In addition, the multi-objective extended compact genetic algorithm~\cite{sastry2005limits} took a marginal product model as the mixture of probability model. Furthermore, a regularity-based model EDA (RM-MEDA) was proposed in~\cite{zhang2008rm} in which the model is built based on the mixture normal distribution over the regularity. Zhou \emph{et al.}~\cite{zhou2009approximating} proposed a regularity-based method for solving the MOPs requiring the objective spaces to be $(m-1)$ dimensions.

It is believed that EDAs are capable of solving MaOPs without suffering the disadvantages of MOEAs with traditional genetic operators. Although, MBN-EDA has shown the promise in solving MaOPs, the development of many-objective optimization EDAs (MaOEDA) is still in infancy. Especially, probability models based on regularity have been extensively investigated in the discipline of statistical learning~\cite{cherkassky2007learning,hastie2005elements}, and regularity-based model are easier to build, and fairly effective. Based on our recent research achievements on this topic~\cite{he2016visualization,cheng2015many} and motivated by the success of regularity-based EDAs for MOPs~\cite{zhou2005model,zhou2006combining,zhang2008rm}, an improved regularity-based EDA for MaOPs, in short named MaOEDA-IR, is proposed in this paper. To be specific, models employed to generate new solutions in the proposed algorithm are built based on a group of neighbors selected by a uniformly distributed \textit{reference vectors}. In order to improve the selection pressure, \textit{diversity repairing} mechanism is developed to prevent the adverse DR phenomenon in each generation and push the new solutions toward having a closer proximity to the PF. Furthermore, \textit{dimension reduction} technique is employed priori to the evolution to reduce the cost for exploration search. Specifically, convergence in the proposed algorithm is guaranteed by repairing diversity and sampling solutions based on the reference vectors, while diversity is facilitated by selecting solutions with the nearest perpendicular distances to the reference vectors. Compared to traditional MaOEAs and EDAs, the contributions of the proposed algorithm are summarized as follows:
\begin{enumerate}
  \item Extend the uses of regularity model-based EDAs to MaOPs. In addition, reference vectors-based diversity mechanism is incorporated into the proposed algorithm to enhance the selection pressure.
  \item Large search space poses a challenge for regularity model-based EDAs as do to all MaOEAs. To this end, dimension reduction technique is utilized in the decision space to speed up the exploration search for sampling promising solutions.
  \item Convergence and diversity are considered equally important in the design of a quality MOEA or MaOEA. In the proposed algorithm, convergence is mainly treated in the first stage (the phase of dimension reduction), while diversity is focused in the second stage.
\end{enumerate}

The remainder of this paper is organized as follows. Related reference vectors-based MaOEA, evolutionary algorithms based on dimension reduction, and the seminal work of regularity-based EDA are reviewed in Section~\ref{section_2}, respectively. In Section~\ref{section_3}, the framework of the proposed algorithm is outlined, and the respective steps are detailed. In addition, the complexity of the proposed algorithm are analyzed, and two crucial sub-components of the proposed algorithm as well as the principles for selecting neighbor solutions for building the model are discussed. In Section~\ref{section_4}, a series of experiments are performed over widely used test suites against chosen peer competitors and the results measured by the selected performance metrics are statistically compared, in addition to experiments on investigating the effect of neighbor size, diversity repairing, and dimension reduction. Finally, conclusion and future work are given in Section~\ref{section_5}.

\section{Related Works}
\label{section_2}
Because the proposed algorithm are mainly concerned on the reference vectors, dimension reduction, and regularity-based EDA model, works related to these aspects are discussed. To be specific, state-of-the-art MaOEA, NSGA-III which is based on the reference vectors, is introduced first. Then the evolutionary algorithms employing dimension reduction are reviewed. Next, RM-MEDA which is a regularity-based EDA is analyzed. Finally, the disadvantages of RM-MEDA for solving MaOPs, and the differences to the proposed algorithm are highlighted.

\subsection{NSGA-III}
The major difference of NSGA-III compared to its predecessor is the diversity improving mechanism which is performed by the reference vectors, and it begins to take effect when $j$ solutions need to be chosen from the non-dominated sorting front $F_i$ where $i>0$, $0<j<|F_i|$, and $|\cdot|$ is a cardinality operator. To be specific, each reference vector is assigned to solutions in $F_k$ where $0<k<i-1$ by calculating which solution has the nearest perpendicular distance to it. Then the reference vector $v$ who is assigned the smallest number of solutions is marked and the solution $p$ in $F_i$ that has the smallest perpendicular distance to $v$ is selected. Next, solution $p$ is removed from $F_i$ and the assigned number of $v$ is increased by $1$. These steps are iterated until $j$ solutions are selected from $F_i$. Because these reference vectors are uniformly distributed, the selected solutions are hopefully evenly distributed in the objective space. Specifically, the reference vectors are uniformly generated in the $R^M_+$ space, and the sum of elements in one reference vector is equal to $1$. However, the problem to be optimized is not necessary in this unit hyperplane. For this purpose, all the objectives are to be normalized priori to calculating the perpendicular distances. This mechanism of uniformly distributed reference vector assisting to select solutions is more and more preferred by MaOEAs due to its explicitly diversity preserving nature.

\subsection{Evolutionary Algorithms Based on Dimension Reduction}
The notion that state-of-the-art MOEAs are capable of solving the MOPs naturally leads to an intent of reducing the number of objective in a MaOP (i.e., dimension reduction in the objective space), and then applying these powerful MOEAs in solving them. Specifically, dimension reduction in the objective space refers to removing the redundant objectives, while the same solutions are obtained as with all objectives involved~\cite{Gal1999Consequences}. With the utilization of dimension reduction, the computational complexity is reduced due to a smaller number of objectives. In summary, the dimension reduction schemes considered in literatures can be sorted into two categories. The first category is often known as the correlation-based methods, such as the works in~\cite{brockhoff2010automated,brockhoff2009objective,lopez2008objective}. In addition, the correntropy principal component analysis (C-PCA), maximum variance unfolding PCA (MVU-PCA)~\cite{saxena2007non}, and PCA-based algorithm~\cite{brockhoff2008handling} were proposed to analyze the correlation between the solutions generated in each generation, while Saxena \textit{et al.}~\cite{saxena2013objective} proposed the linear PCA and nonlinear MVC over a set of Pareto-optimal solutions to check the correlation. In addition, Guo \textit{et al.}~\cite{guo2012new} employed the interdependence factors to identify the correlation for dimension reduction. Recently, Wang and Yao proposed a novel approach to reduce the objective dimension by measuring the linear and nonlinear correlation between objectives using nonlinear correlation information entropy~\cite{Wang2016Objective}. The second category covers the algorithms which employ the dimension reduction based on dominance structure, such as the work~\cite{brockhoff2009objective} in which the $\epsilon$-dominance was employed to identify the redundant objectives. In addition, the Pareto Corner Search Evolutionary Algorithm~\cite{singh2011pareto} utilized the corner sorting technique to find the corner solutions in which the dimension reduction was performed. In summary, the dimension reduction techniques in these algorithms are performed in the objective space, while the proposed improved regularity-based MaOEDA builds its model in the decision space. Moreover, it is common that the number of decision variables is much greater than that of the objectives. As a consequence, it is reasonable to reduce the dimension of the decision space for the purpose of computational efficiency and effectiveness.

\subsection{RM-MEDA}
\begin{figure*}[htp]
\begin{center}
\subfloat[]{\includegraphics[width=0.6\columnwidth]{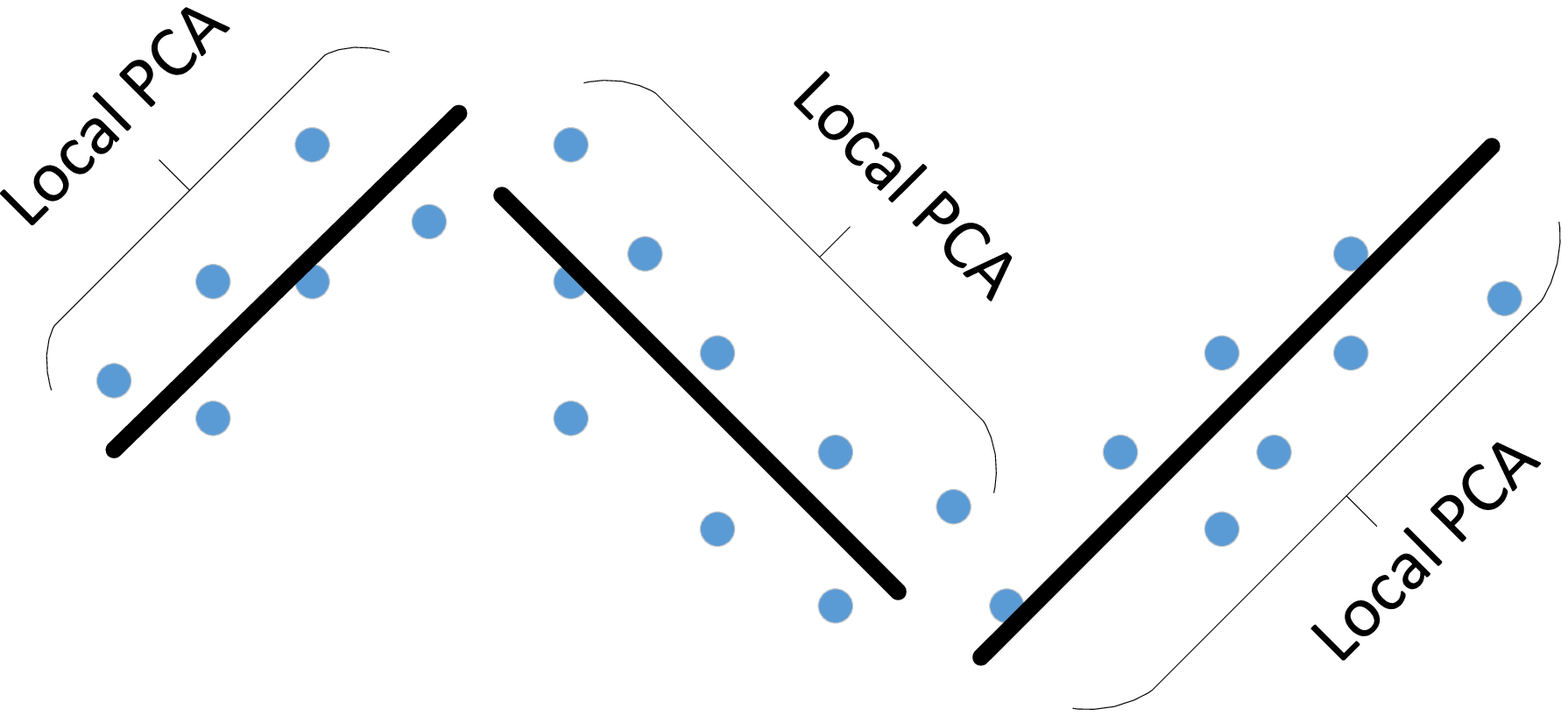}%
\label{fig_local_pca}}
\hfil
\subfloat[]{\includegraphics[width=0.6\columnwidth]{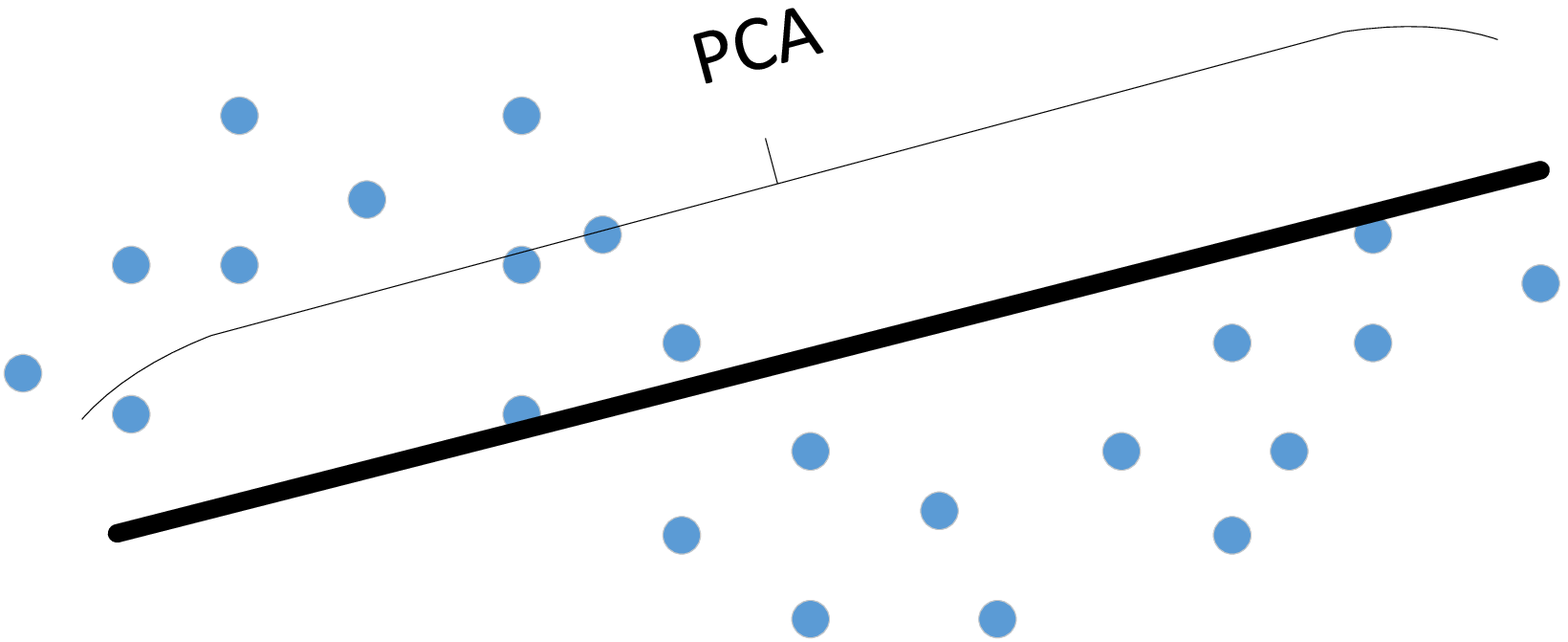}%
\label{fig_pca}}
\caption{An example comparison of local PCA and PCA on the same data. Specifically, Fig.~\ref{fig_local_pca} denotes the utilization of local PCA, while Fig.~\ref{fig_pca} denotes that of PCA. The solid lines on both figures denote the main directions of the principal components. Obviously, local PCA is better to capture the global structure of the give data.}
\label{fig_local_pca_pca}
\end{center}
\end{figure*}

The probability model of RM-MEDA is built based on the regularity of decision space. To be specific, all the solutions are clustered into multiple disjoint groups by local PCA~\cite{kambhatla1997dimension} first, and then models are constructed to generate new solutions for each group. Specifically, local PCA is employed for manifold dimension reduction by performing multiple PCA operations in each piecewise linear segment over the entire given data. Compared to PCA, local PCA is considered better to collectively capture the global structure. For intuitively comparing the effects of local PCA and PCA, an example is plotted in Fig.~\ref{fig_local_pca_pca} in which it is clearly shown that the local PCA works better in estimating the entire structure of the data. Supposed that $\lambda_1,\cdots,\lambda_N$ are the eigenvalues of the covariance matrix of one group, $\lambda_1\geq\cdots\geq \lambda_N$, and the corresponding eigenvectors are $v_1, \cdots, v_N$. Then the model is formulated with the assumption that the PS is a $(M-1)$-dimensional piecewise manifold in a continuous problem. In further, solutions are sampled from this model with the number which is proportional to the volume of the model to that of all the models. Combined with non-dominated sorting, a set of solutions with diversity approximating the PF are generated.

EDAs generate new solutions with probabilistic model to reject the detrimental consequence lead by genetic operator. Especially, regularity-based model is much preferable than most Bayesian network-based models in EDAs because of the simplicity yet effectiveness~\cite{wang2015regularity}. For example, EDAs based on Bayesian network need the procedure of training while regularity-based methods are analytic. However, because RM-MEDA is originally developed for solving MOPs with variables linkage~\cite{zhang2008rm}, it may not be suitable for solving MaOPs. For example, the diversity in RM-MEDA is maintained by uniformly sampling new solutions in the decision space and this cannot give rise to the corresponding diversity in the objective space especially in MaOPs such as all the test problems in DTLZ~\cite{deb2005scalable}. Furthermore, local PCA makes sense when the PS is in full rank in the decision space, which is not necessary the case in MaOPs. To this end, an improved regularity-based EDA for effectively addressing MaOPs is proposed in this paper. In addition, a new diversity facilitating mechanism which employs the uniformly distributed reference vectors is also incorporated to improve the selection pressure. Furthermore, the dimension reduction technique based on the correlation scheme is utilized in the decision space over a set of Pareto-optimal solutions to save the computational complexity. Compared to RM-MEDA, the main contributions of the proposed algorithm are listed as follows.
\begin{enumerate}
  \item The diversity improving mechanism of the proposed algorithm is compatible to most problems while RM-MEDA is only suitable to the problems in which the PS has the same image of PF.
  \item The PS in RM-MEDA is not allowed to lie on any subspaces of the decision space, while the proposed algorithm has no such a requirement of it.
  \item RM-MEDA is implemented with the manifold assumption that the PS is a piece-wise manifold with $(M-1)$-dimension. In this paper, the proposed algorithm is developed without such an assumption.
\end{enumerate}

\section{Proposed Algorithm}
\label{section_3}
In this section, the framework of the proposed algorithm, i.e., improved regularity model-based EDA for many-objective optimization~(MaOEDA-IR), is given first. Then the details of each step in the framework are presented. This is followed by the analysis of the computational complexity. Finally, significant sub-components of the proposed algorithm, and the principles of selecting neighbor solutions are discussed, respectively. It is noted here that the proposed algorithm is given in the context of problem described by Equation~(\ref{equation_mops}).
\subsection{Framework of the Proposed Algorithm}
\label{section3-1}
The proposed algorithm begins with reducing the dimension of the decision space (Subsection~\ref{sec_dimension_reduction}). Then new promising solutions are generated in the reduced decision space to speed up the evolutionary progress and lower the computational cost of exploitation search in the proposed algorithm, and their fitness are evaluated (Subsection~\ref{sec_create_evaluate}). Next, a set of generated reference vectors in $R^M_+$ (it refers to the sub-part of $R^M$, where all points in this sub-part are with element values no less than 0) is mapped (Subsection~\ref{sec_map_reference_vectors}), and regularity-based model is built (Subsection~\ref{sec_model_build}) for repairing the diversity of the proposed algorithm (Subsection~\ref{sec_diversity_repair}) in which a set of solutions $R_t$ are generated. Based on $R_t$, new offspring are generated (Subsection~\ref{sec_generate_offspring}). With the help of environmental selection operator (Subsection~\ref{sec_environmental_selection}), a set of solutions with a better quality in convergence and diversity are obtained. Repairing the diversity, model updating, sampling new solutions, and environmental selection are performed one by one in a limit number of generations. At last, final selection is utilized to choose the representative solutions for the available slots (Subsection~\ref{sec_environmental_selection}). In addition, maximum generation number, a set of uniformly distributed reference vectors, neighbor size, and respective threshold for dimension reduction and model building need to be made available prior to the proposed algorithm running. In summary, the framework of the proposed algorithm is listed in Algorithm~\ref{alg_ireda}.
\begin{algorithm}
  \caption{Main Framework of MaOEDA-IR}
  \label{alg_ireda}
  \KwIn{the maximum number of generations $t_{max}$, a set of unit reference vectors $\textbf{r}_0=\{r_{0,1},\cdots,r_{0,N}\}$, neighbor size $T$, dimension reduction threshold $\alpha$, regularity-based model threshold $\beta$; }
  \KwOut{final population $P$;}
  Reduce the dimension of decision space from $R^n$ to $R^k$\;
  \label{alg_framework_1}
  Create population $P_0$ with $TN$ individuals from $R^k$\;
  Evaluate the fitness of $P_0$\;
  Map reference vectors $\textbf{r}_0$ to $\textbf{v}_0=\{v_{0,1},\cdots,v_{0,N}\}$\;
  Build the regularity-based model $\Phi$\;
  \label{alg_framework_2}
  $t\leftarrow 0$\;
  \While{$t<t_{max}$}
  {
    $R_t\leftarrow$ Repair the diversity\;
    \label{alg_framework_3}
    $S\leftarrow$ Non-dominated selection from $P_t\cup R_t$  \label{alg_add_1}\;

    Map reference vectors $\textbf{v}_{t-1}$ to $\textbf{v}_t=\{v_{t,1}\cdots,v_{t,N}\}$ \label{alg_add_2}\;

    $Q_t\leftarrow$ Generate offspring\;
    $S\leftarrow$ Non-dominated selection from $P_t\cup R_t \cup Q_t$\label{alg_add_3}\;

    Update reference vectors $\textbf{v}_t=\{v_{t,1}\cdots,v_{t,N}\}$ \label{alg_add_4}\;

    $P_{t+1}\leftarrow$ Environmental selection from $Q_t\cup R_t \cup P_{t}$\;
    \label{alg_framework_4}
    $t\leftarrow$ $t+1$\;
  }
  $P\leftarrow$ Final selection from $P_{t_{max}}$\;
  \textbf{Return} $P$.
\end{algorithm}

\subsection{Reducing the Dimension of Decision Space}
\label{sec_dimension_reduction}
Dimension reduction technique is used to reduce the volume of exploration space to speed up the search of sampling new solutions. Ideally, the subspace of PS is desirable. To this end, a set of Pareto-optimal solutions is suitable to be the training data and then exploitation is performed in the subspace. Noted here that, the training data only require the solutions with convergence and the diversity is not necessary. Consequently, algorithms such as the conventional weighted aggregation method~\cite{fleming85computer} for problems with convex PF, and the evolutionary dynamic weighted aggregation methods~\cite{jin2001adapting} for problems with non-convex PF are suitable for this. In this paper, the Pareto Corner Search Evolutionary Algorithm (PCSEA)~\cite{singh2011pareto} is employed because 1) a set of corner Pareto-optimal solutions which can be used as the training data as well as the extreme points (extreme points are employed for building the regularity-based model in Subsection~\ref{sec_model_build}) are obtained simultaneously when the PCSEA completes, 2) the source code is available, and 3) the computational cost is promising compared to the algorithms selected for generating the extreme points and the training data. Moreover, the PCSEA has been successfully employed to generate solutions for dimension reduction in the objective space in its seminal paper. In each generation of PCSEA, $2M$ lists are increasingly ordered, where $M$ is the number of objectives. Specifically, the first $M$ lists are about the $M$ objectives, while the last $M$ lists are with the exclusive $L_2$ norm square. Especially, the exclusive $L_2$ norm square of the $i$-th objective is with the form $\sum_{j=1,j\neq i}^Mf_i^2$. By assigning the solutions in the top of the lists with a smaller rank value, the corner solutions are highlighted (more details can be found in~\cite{singh2011pareto}). When the evolution process of PCSEA is completed, the Pareto-optimal solutions, which are denoted by $S_p$, are selected from the population.

For convenience of the development, a matrix $X$ is used to represent $S_p$. Specifically, each row in $X$ denotes one solution while the columns refer to the different dimension of decision variables. Then the process of dimension reduction is illustrated in Algorithm~\ref{alg_dimension_reduction} in which it can be divided into two parts. The first part is to employ the principal component analysis (PCA) to find the projected space in which $X$ keeps its $(\alpha\times 100\%)$ information where $\alpha$ is a predefined threshold (lines~\ref{alg_dimension_reduction_begin_pca}-\ref{alg_dimension_reduction_end_pca}). After the transformed data is projected back to its original space (in line~\ref{alg_dimension_reduction_restore}), the values at reduced dimensions become $0$. Thereafter, the index of the reduced dimensions $I$ are selected, which are implemented in lines~\ref{alg_dimension_reduction_saved_begin}-\ref{alg_dimension_reduction_saved_end}. Then, $I$ is saved together with the mean value $\mu$ of $X$ for evaluating the fitness of solutions generated in the reduced decision variable space.

\begin{algorithm}
  \caption{Dimension Reduction}
  \label{alg_dimension_reduction}
    \KwIn{the matrix $X$ representing $S_p$, threshold $\alpha$;}
    \KwOut{the index of the removed dimension, the mean value of $X$;}
    $\mu \leftarrow$ Compute the column mean value of $X$\;
    \label{alg_dimension_reduction_begin_pca}
    $X'\leftarrow $ Subtract $\mu$ from $X$\;
    $U\leftarrow $ Select principal components with threshold $\alpha$\;
    \label{alg_dimension_reduction_end_pca}
    $\hat{X}\leftarrow U'UX'$\;
    \label{alg_dimension_reduction_restore}
    $I\leftarrow\emptyset; V\leftarrow\emptyset$\;
    \label{alg_dimension_reduction_saved_begin}
    \For{$j\leftarrow 1$ \rm{\textbf{to}} $n$}
    {
        $s\leftarrow$ compute the $j$-th column sum of $\hat{X}$\;
        \If{ $s == 0$}
        {
            $I\leftarrow I\cup j$\;
        }
    }
    \label{alg_dimension_reduction_saved_end}
    \textbf{Return} $I$, $\mu$.
\end{algorithm}

In particular, the main reason for not using $U$ but the original space is for reducing the computational complexity. Specifically, $U$ is not the original decision space, and the solutions sampled from $U$ cannot be directly used for fitness evaluation. There are two ways to solve this problem. The first one is to sample solutions from $U$ and then transform solutions into the original space when they are evaluated for the fitness. The other one is to transform $U$ to the original space in advance, and then sample solutions from the original space. If we use the first method, the transformation operation needs to be performed for each solution in each generation. However, if we use the second method, we only need to do this transformation once. Obviously, the second method is with less computational complexity. Next, we will explain the mechanism of lines~\ref{alg_dimension_reduction_saved_begin}-\ref{alg_dimension_reduction_saved_end} in Algorithm~\ref{alg_dimension_reduction}.

By using PCA, the generated principal space $U$ is with less number of dimensions compared to that of the original space. When $U$ is transformed to the original space, the values of the dimensions reduced by PCA are all zeros. Therefore, we use lines~\ref{alg_dimension_reduction_saved_begin}-\ref{alg_dimension_reduction_saved_end} in Algorithm~\ref{alg_dimension_reduction} to find these dimensions by checking where their values are zeros. Once we find these dimensions, we remove them and store their indexes. In Algorithm~\ref{alg_dimension_reduction} we use $I$ to denote the space without the reduced dimensions, and sample solutions from $I$. When these solutions are used for fitness evaluation, just padding their corresponding mean value (stored in $\mu$) to the corresponding dimension based on the stored index, and using them to do fitness evaluation.

In most PCA-based methods, none of solutions sampled from the reduced space needs to be operated in the original space. However, in the proposed design, the solutions must be transformed back to the original space for fitness evaluation. Naturally, the dimension reduction technique used here is quite different from a general PCA-based method.
\subsection{Creating Population and Evaluating Fitness}
\label{sec_create_evaluate}
Assuming the dimension of the reduced decision variable space is $k$ (obviously $k=n-|I|$). Based on the design principles of the proposed algorithm, each final solution of the proposed algorithm is desirable to be associated with one reference vector, and the model is built on the neighbor solutions of one particular reference vector. Therefore, the population is randomly initialized in $R^k$ with size $TN$. In order to evaluate the fitness, the created population needs to be translated back to $R^n$, which is demonstrated by Algorithm~\ref{alg_translate_pop}. Specifically, the translated population $P_0'$ has the same number to that of the initialized population $P_0$, and each individual in $P_0'$ is with $n$-dimension. Furthermore, the values in reduced dimension are equal to that of the elements in the mean value vector $\mu$. To this end, $P_0'$ is initialized in $R^{TN\times n}$ (line~\ref{alg_translate_pop_1}) with zeros first. Then, the element values in each row of $P_0'$ is set to be as $\mu$ (line~\ref{alg_translate_pop_2}). Finally, each column of $P_0$ is added to the corresponding column of $P_0'$, which is implemented by lines~\ref{alg_translate_pop_3}-\ref{alg_translate_pop_4}. Once the translation is performed, the fitness is evaluated by introducing the population to the problem to be optimized.
 \begin{algorithm}
  \caption{Translate the Population}
  \label{alg_translate_pop}
    \KwIn{population $P_0\in R^k$, index of reduced dimension $I$, mean value $\mu$;}
    \KwOut{population $P_0'\in R^n$;}
    Initialize a matrix $P_0'\in R^{TN\times n}$ with zeros\;
    \label{alg_translate_pop_1}
    Copy $\mu$ to each row of $P_0'$\;
    \label{alg_translate_pop_2}
    $l\leftarrow$ $1$\;
    \For{$i\leftarrow$ $1$ \rm{\textbf{to}} $n$ and $i$ \rm{\textbf{not in}} $I$}
    {
        \label{alg_translate_pop_3}
        Add the $l$-th column of $P_0$ to the $i$-th column of $P_0'$\;
        $l\leftarrow l+1$\;
    }
    \label{alg_translate_pop_4}
    \textbf{Return} $P_0'$.
\end{algorithm}
\subsection{Mapping Reference Vectors}
\label{sec_map_reference_vectors}
\begin{figure*}[htp]
\begin{center}
\subfloat[]{\includegraphics[width=0.5\columnwidth]{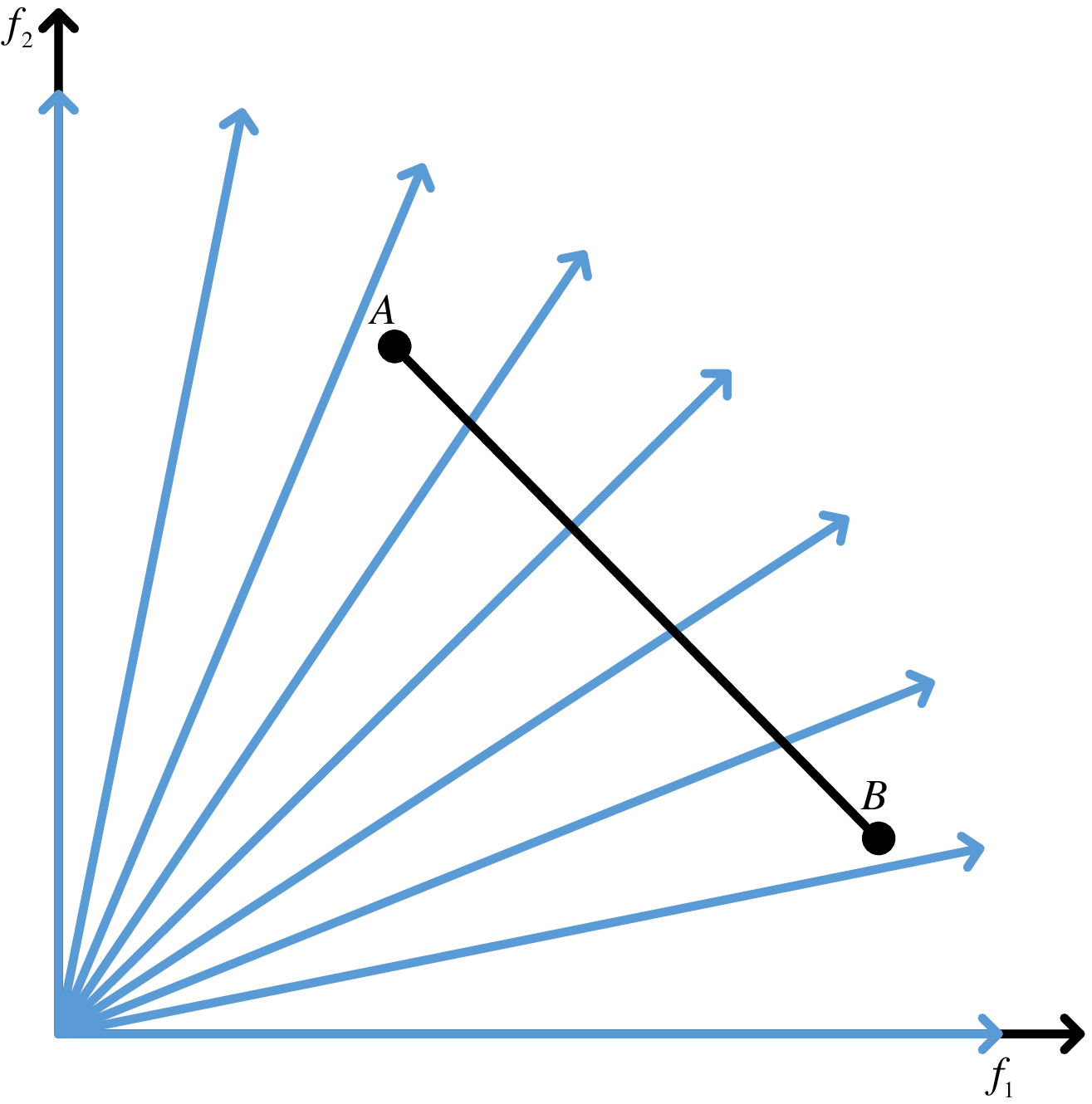}%
\label{fig_mapping_before}}
\hfil
\subfloat[]{\includegraphics[width=0.5\columnwidth]{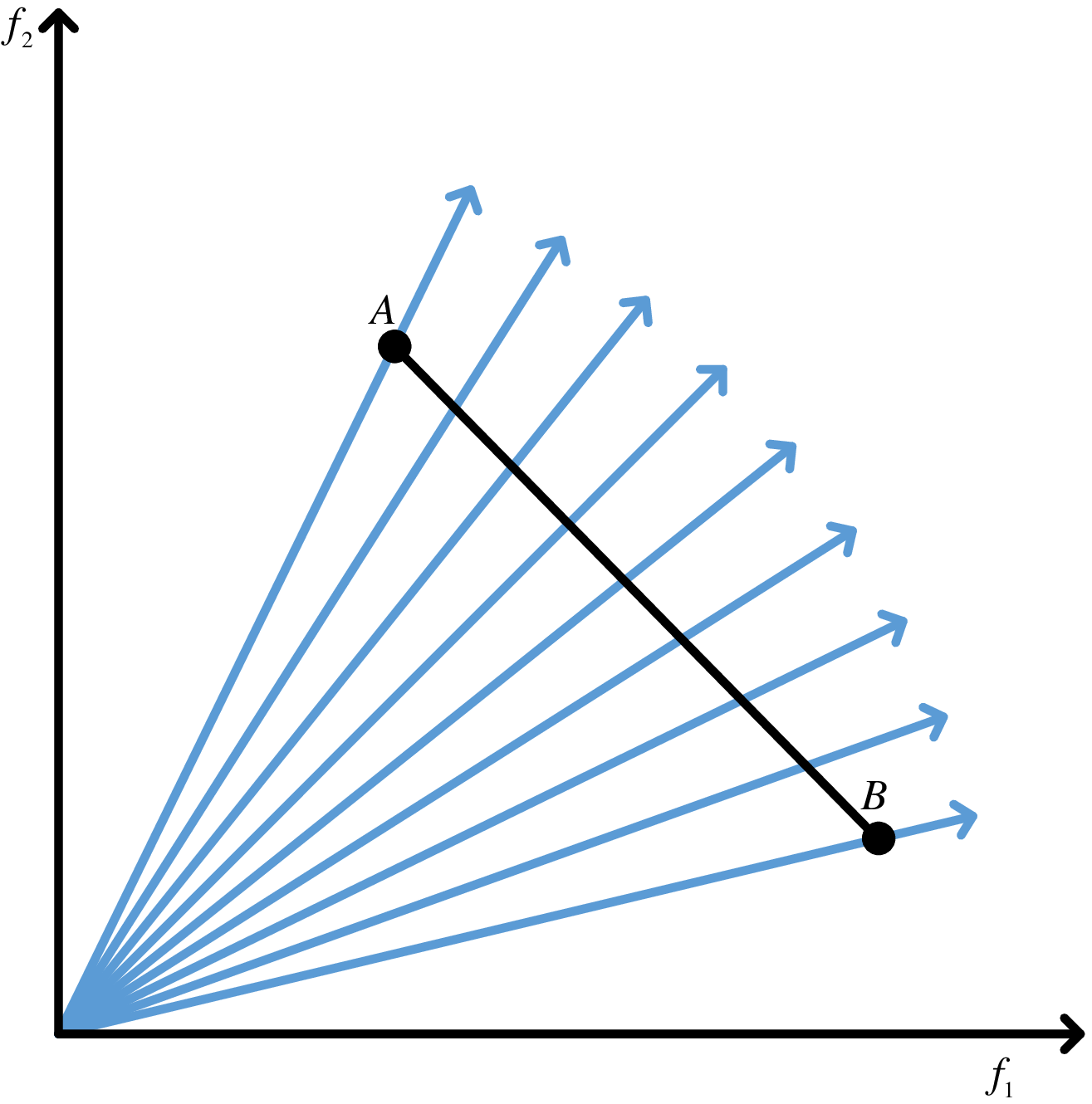}%
\label{fig_mapping_after}}
\hfil
\subfloat[]{\includegraphics[width=0.5\columnwidth]{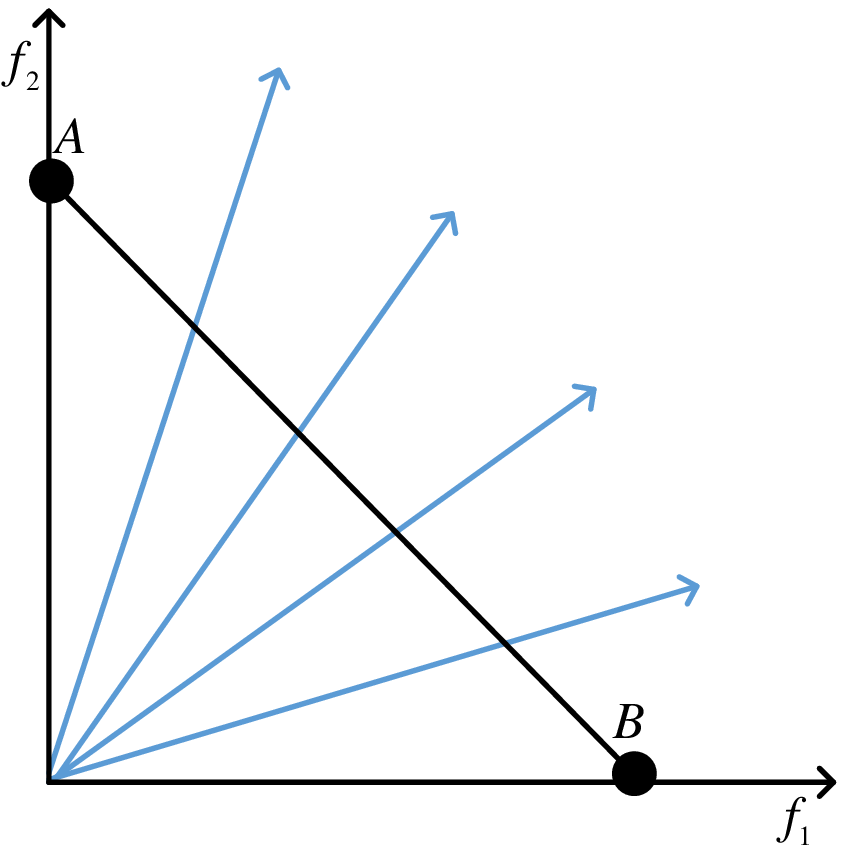}%
\label{fig_no_mapping}}
\caption{A $2$-dimensional example to illustrate the motivation of mapping the reference vectors. Specifically, there are nine reference vectors (blue lines) generated by the Das and Dennis's method, and line $AB$ denotes the Pareto front. The reference vectors without mapping are plotted in Fig.~\ref{fig_mapping_before} in which only four reference vectors intersect $AB$, while the reference vectors which have been mapped are plotted in Fig.~\ref{fig_mapping_after} in which all the sampled reference vectors intersect $AB$ (i.e., the situation that Algorithm~\ref{alg_mapping_reference_vectors} have been performed). Fig.~\ref{fig_no_mapping} denotes the situation that only lines~\ref{alg_begin_extreme_points}-\ref{alg_begin_intercepts} in Algorithm~\ref{alg_mapping_reference_vectors} have been performed.}
\label{fig_motivation_map_rf}
\end{center}
\end{figure*}

\begin{algorithm}
  \caption{Mapping Reference Vectors}
  \label{alg_mapping_reference_vectors}
    \KwIn{reference vectors $\textbf{r}_0$, data $S_p$ from Algorithm~\ref{alg_dimension_reduction};}
    \KwOut{mapped reference vector $\textbf{v}_0$;}
    $F\leftarrow$ Calculate the objective values of $S_p$\;
    \label{alg_begin_extreme_points}
    $\textbf{z}^\textbf{u}\leftarrow$ $\emptyset$ \;
    \For{$i\leftarrow 1$ \rm{\textbf{to}} $M$}
    {   $\textbf{v}\leftarrow[0,\cdots,0]$\;
        \For{\rm{\textbf{each}} $f$ \rm{\textbf{in}} $F$}
        {
            \If{$f_i>\textbf{v}_i$}
            {
                $\textbf{v}\leftarrow f$\;
            }
        }
        $\textbf{z}^\textbf{u}\leftarrow \textbf{z}^\textbf{u} \cup \textbf{v}$\;
    }
    \label{alg_end_extreme_points}
    $\textbf{z}^*\leftarrow\emptyset$\;
    \label{alg_begin_ideal_points}
    \For{\rm{\textbf{each}} $f$ \rm{\textbf{in}} $F$}
    {
        $v\leftarrow +\infty$\;
        \For{$i\leftarrow 1$ \rm{\textbf{to}} $M$}
        {
            \If{$f_i<v$}
            {
                $v\leftarrow f_i$\;
            }
        }
        $\textbf{z}^*\leftarrow \textbf{z}^*\cup v$\;
    }
    \label{alg_end_ideal_points}

    Update each $z$ in $\textbf{z}^\textbf{u}$ by $z\leftarrow z - \textbf{z}^*$\;
    \label{alg_begin_intercepts}
    $\textbf{a}=[a_1\cdots,a_M]\leftarrow$ Find the intercepts of the hyperplane constructed by $\textbf{z}^\textbf{u}$\;
     \label{alg_end_intercepts}
    \For{$i\leftarrow 1$ \rm{\textbf{to}} $N$}
    {
    \label{alg_begin_mapping_rv}
        $v_{0,i}\leftarrow$ = $r_{0,i}\times \textbf{a} + \textbf{z}^*$\;
    }
    \label{alg_end_mapping_rv}
    \textbf{Return} $\textbf{v}_0=\{v_{0,1},\cdots,v_{0,N}\}$.
\end{algorithm}
Conventionally, Das and Dennis's method~\cite{das1998normal} is employed to generate the uniformly distributed reference vectors $\textbf{r}_0$ which is constructed in $R^M_+$. However, the PF of the problem to be optimized does not necessary span the entire $R^M_+$ space. In order to keep the same number of reference vectors in the PF, $\textbf{r}_0$ needs to be mapped. For illustrating this motivation, an example in the $2$-dimensional space is plotted in Fig.~\ref{fig_motivation_map_rf} in which nine blue lines denote the reference vectors and line $AB$ denotes the PF. Specifically, Fig.~\ref{fig_mapping_before} describes the reference vectors generated by Das and Dennis's method and only four reference vectors intersect $AB$, while Fig.~\ref{fig_mapping_after} depicts the nine reference vectors intersect $AB$ after they are mapped. Because the final solutions in the proposed algorithm are selected by their corresponding reference vector, it is obvious that the design in Fig.~\ref{fig_mapping_after} is preferred. In addition, Algorithm~\ref{alg_mapping_reference_vectors} presents the details of mapping the reference vectors which can be divided into four steps. First, the objective values of the training data for dimension reduction in Algorithm~\ref{alg_dimension_reduction} and the extreme points (denoted by $\textbf{z}^\textbf{u}$) are calculated (lines~\ref{alg_begin_extreme_points}-\ref{alg_end_extreme_points}). Second, the ideal point is derived by selecting the minimum values in all objectives, which are implemented in lines~\ref{alg_begin_ideal_points}-\ref{alg_end_ideal_points}. Noted here that, this approach to calculate the ideal point is also utilized in~\cite{yuan2015balancing,cheng2016reference,deb2009hybrid,deb2006towards,wang2015nadir}. Third, the extreme points are updated by subtracting the ideal point and the intercepts are calculated (lines~\ref{alg_begin_intercepts}-\ref{alg_end_intercepts}) by solving the equation\footnote{Repetitive points existing in $\textbf{z}^\textbf{u}$ will lead multiple solutions to this equation. To avoid this, the trick introduced in~\cite{yuan2014evolutionary} is employed. $\textbf{z}^\textbf{u}\times (1/a)=I$ where $I$ is an identity vector. Fourth, the reference vectors uniformly generated in $R^M_+$ are mapped by lines~\ref{alg_begin_mapping_rv}-\ref{alg_end_mapping_rv} as mapped reference vectors $\textbf{v}_0$. Noting that, the last two steps are necessary. Otherwise, only the origin of the coordinate system has been moved to the idea point, and there is still no sufficient numbers of reference vectors intersecting with line $AB$ (see Fig.~\ref{fig_no_mapping}).

}

\subsection{Building the Regularity-based Model}
\label{sec_model_build}
Generally, the regularity-based model is composed of multiple sub-models due to the complexity of regularity on which a unified model is difficult to exactly capture the global intrinsic relation~\cite{zhang2008rm,zhang2006modelling}. In this proposed algorithm, each sub-model is built based on one reference vector with its neighbor solutions and Algorithm~\ref{alg_model_build} presents the details.
\begin{algorithm}
  \caption{Build the $j$-th Regularity-based Sub-model}
  \label{alg_model_build}
    \KwIn{neighbor solutions $S_n$ of the $j$-th reference vector, threshold $\beta$, enlargement factor $\gamma$;}
    \KwOut{model $\Phi$;}
    Let matrix $X$ denote $S_n$\;
    \label{alg_model_build_1_begin}
     $\mu \leftarrow$ Compute the column mean value of $X$\;
    $X'\leftarrow$ Subtract $\mu$ from each rom in $X$\;
    \label{alg_model_build_1_end}
    $[\lambda_1,\cdots,\lambda_M],[v_1,\cdots,v_M]\leftarrow$ Eigen-factorize the covariance matrix of $X$ and descend the eigenvalues and eigenvectors\;
    \label{alg_model_build_2}
    $i\leftarrow 0$\;
    \label{alg_model_build_3_begin}
    \While{$(\lambda_1+\cdots+\lambda_i)/(\lambda_1+\cdots+\lambda_M) < \beta$}
    {
        $i\leftarrow i + 1$\;
    }
    \label{alg_model_build_3_end}
    $y\leftarrow X'\times [v_1,\cdots,v_i]$\;
    \label{alg_model_build_4}
    $l,u\leftarrow$ Find the minimum and the maximum values in each row of $y$\;
    \label{alg_model_build_5}
    $\Omega\leftarrow \left\{\mu+\sum_{j=1}^i\tau v_i, l-\gamma(u-l)\leq \tau \leq u+\gamma(u-l)\right\}$\;
    \label{alg_model_build_6}
    $\epsilon \leftarrow \frac{1}{M-i+1}\sum_{j=i}^M\lambda_j$\;
    \label{alg_model_build_7}
    \textbf{Return} $\Phi=\Omega+\epsilon$.
\end{algorithm}

To be specific, given the neighbor solutions $S_n$ of the $j$-th reference vector, the threshold $\beta$, and the enlargement factor $\gamma$. The steps of building the $j$-th sub-model are illustrated as follows. First, $S_n$ is represented by the matrix $X$ and centered (lines~\ref{alg_model_build_1_begin}-\ref{alg_model_build_1_end}). Then the eigenvalues as well as the eigenvectors of $X$ are obtained and ordered based on descending the eigenvalues (line~\ref{alg_model_build_2}). Lines~\ref{alg_model_build_3_begin}-\ref{alg_model_build_3_end} demonstrates the search of principal components on which the centered data is projected (line~\ref{alg_model_build_4}). Next, the projected space is constrained by find the minimum and the maximum values of projection (line~\ref{alg_model_build_5}), and the latent space for generating new offspring is obtained with the enlargement factor and the principle components (line~\ref{alg_model_build_6}). Finally, the noise for the latent space is computed from the mean values of the eigenvalues regarding the non-principle components (line~\ref{alg_model_build_7}), and the regularity-based model is obtained.

\subsection{Repairing the Diversity}
\label{sec_diversity_repair}
Diversity repairing is employed by sampling new solutions to mitigate the adverse of the phenomenon that reference vectors lack associated solutions. To this end, all the non-dominated individuals in the current population are enumerated to find their respective nearest reference vectors first, which is implemented by lines~\ref{alg_repair_diversity_begin_1}-\ref{alg_repair_diversity_end_1}. Then lines~\ref{alg_repair_diversity_begin_2}-\ref{alg_repair_diversity_end_2} demonstrate the selection of neighbor solutions. In addition, the model building and new solution sampling are presented in line~\ref{alg_repair_diversity_3} and lines~\ref{alg_repair_diversity_begin_4}-\ref{alg_repair_diversity_end_4}, respectively. Noted in the phase of selecting neighbor solutions for reference vector $v_{t,i}$, its neighbor solutions are from the current population $S_t$ and the non-dominated solutions in $S_t$, the motivation of which is discussed in Subsection~\ref{sec_discussion}. In addition, it is obvious that the size of the neighbor solutions is not necessary with $T$. I.e., it is with $T+1$ when the selected non-dominated neighbor solution has been included in line~\ref{alg_repair_diversity_begin_2}. The reasons that the neighbor solution size are not strictly kept with $T$ are that 1) it does not affect the built model and 2) removing the extra solution gives more computational cost.

\begin{algorithm}
  \caption{Repair the Diversity}
  \label{alg_repair_diversity}
    \KwIn{current population $P_t$, reference vectors $\textbf{v}_t$, neighbour size $T$;}
    \KwOut{new solutions $R_t$;}
    $S=\{s_1,\cdots,s_k\}\leftarrow$ Non-dominated selection from $P_t$\;
    $I\leftarrow \emptyset$\;
    \label{alg_repair_diversity_begin_1}
    \For{\rm\textbf{each} $s$ \rm\textbf{in} $S$}
    {
        $I\leftarrow I\cup\argmin_{i} ||s-v_{t,i}sv_{t,i}^T/(v_{t,i}v_{t,i}^T)||$\;
    }
    \label{alg_repair_diversity_end_1}
    $R_t\leftarrow \emptyset$\;
    \For{$i\leftarrow 1$ \rm\textbf{to} $N$ \rm\textbf{and} $i$ \rm\textbf{not in} $I$}
    {
        $\text{neighbour}(i)\leftarrow$ Select $T$ solutions from current population who have the smallest perpendicular distances to $v_{t,i}$\;
        \label{alg_repair_diversity_begin_2}
        $j\leftarrow \argmin_{j} ||s_j-v_{t,i}s_jv_{t,i}^T/(v_{t,i}v_{t,i}^T)||$\;
        $\text{neighbour}(i)\leftarrow \text{neighbour}(i)\cup s_j$\;
        \label{alg_repair_diversity_end_2}
        Build model $\Phi_i=\Omega_i+\epsilon_i$ with $\text{neighbour}(i)$\;
        \label{alg_repair_diversity_3}
        $A\leftarrow$ Uniformly sample $T$ points in $[l-\gamma(u-l), u+\gamma(u-l)]$ from $\Omega_i$\;
        \label{alg_repair_diversity_begin_4}
        $B\leftarrow$ Sample points from the normal distribution with mean $0$ and standard derivation $\sqrt{\epsilon_i}$\;
        $R\leftarrow (A+B)$\;
        \label{alg_repair_diversity_end_4}
        $R_t\leftarrow R_t\cup R$\;
    }
    \textbf{Return} $R_t$.
\end{algorithm}

\subsection{Generating Offspring}
\label{sec_generate_offspring}
After repairing the diversity, the solution set $R_t$ is generated. Then, non-dominated solutions in $S$ are selected from $R_t$ and the current population $P_t$ (line~\ref{alg_add_1} of Algorithm~\ref{alg_ireda}). Next, reference vectors $\textbf{v}_t=\{v_{t,1}, \cdots, v_{t,N}\}$ are mapped (line~\ref{alg_add_2} of Algorithm~\ref{alg_ireda}), which is performed by Algorithm~\ref{alg_mapping_reference_vectors} with the input parameters $\textbf{v}_{t-1}$ and $S$. Finally, offspring $Q_t$ are generated by Algorithm~\ref{alg_offspring_generate}. In summary, generating offspring can be viewed as the diversity repairing in each reference vector, which could be investigated through the analogy between lines~\ref{alg8_begin}-\ref{alg8_end} of Algorithm~\ref{alg_offspring_generate} and lines~\ref{alg_repair_diversity_begin_2}-\ref{alg_repair_diversity_end_4} of Algorithm~\ref{alg_repair_diversity}. However, the motivation of generating offspring is different to that of diversity repairing, which will be discussed in Subsection~\ref{sec_discussion}. In addition, updating the reference vectors is motivated by achieving a better performance of the proposed algorithm, although it has been reported that PCSEA is capable of finding the approximated corner solutions from which the extreme points are derived~\cite{singh2011pareto}.

\begin{algorithm}
  \caption{Generate Offspring}
  \label{alg_offspring_generate}
    \KwIn{non-dominated solutions $S=\{s_1,\cdots,s_k\}$,} neighbour size $T$, reference vectors $\textbf{v}_{t}$;
    \KwOut{offspring $Q_t$;}

    $Q_t\leftarrow \emptyset$\;
    \For{$i\leftarrow 1$ \rm\textbf{to} $N$}
    {
        $\text{neighbour}(i)\leftarrow$ Select $T$ solutions from $S$ who have the smallest perpendicular distances to $v_{t,i}$\;
        \label{alg8_begin}
        $j\leftarrow \argmin_{j} ||s_j-v_{t,i}s_jv_{t,i}^T/(v_{t,i}v_{t,i}^T)||$\;
        $\text{neighbour}(i)\leftarrow \text{neighbour}(i)\cup s_j$\;
        Build model $\Phi_i=\Omega_i+\epsilon_i$ with $\text{neighbour}(i)$\;
        $A\leftarrow$ Uniformly sample $T$ points in $[l-\gamma(u-l), u+\gamma(u-l)]$ from $\Omega_i$\;
        $B\leftarrow$ Sample points from the normal distribution with mean $0$ and standard derivation $\sqrt{\epsilon_i}$\;
        $Q\leftarrow (A+B)$\;
        \label{alg8_end}
        $Q_t\leftarrow Q_t\cup Q$\;
    }
    \textbf{Return} $Q_t$.
\end{algorithm}

\subsection{Environmental Selection and Final Selection}
\label{sec_environmental_selection}

After offspring $Q_t$ are generated, non-dominated solutions $S$ are selected from the current population, i.e., $P_t\cup R_t\cup Q_t$ (line~\ref{alg_add_3} of Algorithm~\ref{alg_ireda}). Then, reference vectors $\textbf{v}_t=\{v_{t,1}, \cdots, v_{t,N}\}$ are updated (line~\ref{alg_add_4} of Algorithm~\ref{alg_ireda}) by performing Algorithm~\ref{alg_mapping_reference_vectors} with the input parameters $S$ and itself. Next, the environmental selection is performed. Specifically, the environmental selection aims at removing ill-fit solutions from the current population and maintaining a limit size of representative individuals to reduce the cost of computation in each generation. The final selection is to choose the best-fit solutions. Moreover, both selections are dependent in the proposed algorithm. In the framework of the proposed algorithm, it is assumed that $N$ solutions are needed by the decision-maker when the algorithm is finished. Because the proposed algorithm is based on the statistics of regularity of the $T$ neighbors regarding each solution in $N$. As a consequence, the number of solutions for building the model should be $TN$. In addition, extra solutions are incorporated from the phases of diversity repairing and offspring generating in each generation. To this end, the purpose of environmental selection is for maintaining a size of population with the same number of initialized population while the final selection is for selecting $N$ solutions. It is hopeful that each reference point has $T$ neighbor solutions after the environmental selection which is described by Algorithm~\ref{alg_environment_selection}. Specifically, final selection is implemented if the number $T$ is replaced by $1$ in the environmental selection.

\begin{algorithm}
  \caption{Environmental Selection}
  \label{alg_environment_selection}
    \KwIn{non-dominated solutions $S=\{s_1,\cdots,s_k\}$,} neighbour size $T$, reference vectors $\textbf{v}_{t}$;
    \KwOut{$P_{t+1}$;}
    $L_1,\cdots,L_N\leftarrow\emptyset$\;
    \label{alg_es_begin_1}
    \For{\rm\textbf{each} $s$ \rm\textbf{in} $S$}
    {
        $i\leftarrow\argmin_{i} ||s-v_{t,i}sv_{t,i}^T/(v_{t,i}v_{t,i}^T)||$\;
        $L_i\leftarrow L_i\cup s$\;
    }
    \label{alg_es_end_1}
    \For{$i\leftarrow 1$ \rm\textbf{to} $N$}
    {
    \label{alg_es_begin_2}
        \If{$|L_i| > T$}
        {
            $\{F_1,\cdots, F_l\}\leftarrow$ Non-dominated sorting solutions in $L_i$\;
            $k\leftarrow 1$, $L_i\leftarrow \emptyset$\;
            \While{$|L_i|+|F_k| \leq T$}
            {
                $L_i\leftarrow L_i\cup F_k$\;
                $k\leftarrow k + 1$\;
            }
            \If{$|L_i|<T$}
            {
                $D\leftarrow$ Select $T-|L_i|$ solutions from $F_k$ who have the smallest perpendicular distances to $v_{t,i}$\;
                $L_i\leftarrow L_i\cup D$\;
            }
        }
    }
    \label{alg_es_end_2}
    From current population removing solutions in $L_i\cup\cdots\cup L_N$\;
    \label{alg_es_4}
    \For{$i\leftarrow 1$ \rm\textbf{to} $N$}
    {
    \label{alg_es_begin_3}
        \uIf{$|L_i| < T$}
        {
            $D\leftarrow$ Select $T-|L_i|$ solutions from current population who have the smallest perpendicular distances to $v_{t,i}$\;
            $L_i\leftarrow L_i\cup D$\;

        }\Else{
            $D\leftarrow$ Select $T$ solutions from $L_i$ who have the smallest perpendicular distances to $v_{t,i}$\;
        }
        From current population removing solutions in $D$\;
        \label{alg_es_5}
    }
    \label{alg_es_end_3}
    \textbf{Return} $Q_t=L_i\cup\cdots\cup L_N$.
\end{algorithm}

In summary, the environmental selection covers two steps. The first step is to assign the reference vectors by selecting the non-dominated solutions which have smallest perpendicular distances to them (lines~\ref{alg_es_begin_1}-\ref{alg_es_end_1} of Algorithm~\ref{alg_environment_selection}). The other is to set $T$ solutions for each reference vector. Specifically, the non-dominated sorting and truncated selection is employed when more than $T$ solutions are assigned to one reference vector (lines~\ref{alg_es_begin_2}-\ref{alg_es_end_2} of Algorithm~\ref{alg_environment_selection}), otherwise necessary solutions are selected from the current population based on the smallest perpendicular distances (lines~\ref{alg_es_begin_3}-\ref{alg_es_end_3} of Algorithm~\ref{alg_environment_selection}). Noted that, selected solutions are removed from the current population to avoid repetitions being re-selected (lines~\ref{alg_es_4} and \ref{alg_es_5} of Algorithm~\ref{alg_environment_selection}).
\subsection{Computational Complexity}
\label{sec_computational_complexity}
Complexity of the proposed algorithm is analyzed with the context of problem defined by Equation~(\ref{equation_mops}). For convenience, the number of solutions for PCSEA to generate the training data for dimension reduction is set to be same to that in the final selection of the proposed algorithm. Consequently, the computational times of PCSEA is $O(MNlog(N))$. For the dimension reduction, the most cost is for computing the eigenvalues and eigenvectors which has $O((TN)^3)$ computations. In summary, the computational times of dimension reduction is $O((TN)^3)$. I addition, the creating population and fitness evaluation need $O(TNk)$ and $O(TNM)$ computation times, respectively. The computational complexity of mapping reference vectors is $O(TNM)$. Moreover, the complexity of building the model is mainly contributed by the matrix factorization whose complexity is $O(T^3)$. Briefly, lines~\ref{alg_framework_1}-\ref{alg_framework_2} in Algorithm~\ref{alg_ireda} takes $O((TN)^3)$ computational times. Furthermore, the model building, non-dominated selection from $P_t\cup R_t$, non-dominated selection from $P_t\cup R_t \cup Q_t$ dominate the computational complexity of the diversity repairing, generating offspring, and environmental selection. As a consequence, the computational complexity of lines~\ref{alg_framework_3}-\ref{alg_framework_4} is $O(MN^2)$ or $O(T^3)$ which is greater. In addition, the final selection takes $O(MN^2)$ computational times. Generally, the neighbor size $T$ is generally set to be a number with order of magnitude $1$, and the maximum generation is set to that of magnitude $1$. In summary, the computational complexity of the proposed algorithm is $O((TN)^3)$, where $T$ is the neighbor size, and $N$ is the number of solutions the decision-makers require.

\subsection{Discussion}
\label{sec_discussion}
In the proposed algorithm, two sub-components, dimension reduction, and diversity repairing are well-designed to guarantee the performance. In this section, their design motivations are discussed first, and then the experimental verification are presented in Sections~\ref{sec_investigation_diversity_raparation} and~\ref{sec_investigation_dimension_reduction}.

First, redundancy exists in the high dimensional data, and data from a small part of the dimensions is sufficient to represent these high dimensional data, such as in the feature selection discipline. In order to obtain these low dimensional data, dimension reduction technique is utilized. Employing these low dimensional data against the original high dimensional data can significantly benefit the utilization of data, such as lowering the computational cost, improving the precision by removing the interference from the elements in the redundant dimensions, and so on. Furthermore, a series of literatures~\cite{brockhoff2010automated,lopez2008objective,brockhoff2008handling,saxena2007non,saxena2013objective,guo2012new,brockhoff2009objective,singh2011pareto} have been proposed to reduce the objective number in MaOPs, thereafter state-of-the-art MOEAs can be utilized to solve them. Generally, the numbers of decision variables are greater than that of objectives in MaOPs, such as the $M$-dimensional DTLZ7~\cite{deb2005scalable} with $19+M$ decision variables, and as well in MOPs, for example, 2-dimensional ZDT1 problem~\cite{zitzler2000comparison} with $30$ decision variables. Actually, it is no surprise that problems are with the number of decision variables greater than that of objectives because it is difficult to determine which particular factors affect the response, and a better way for modeling is to select all the observed factors, which is always with a larger size. As a result, dimension reduction in the decision space is appropriate and well justified, specifically for the proposed algorithm which is based on the regularity of the decision space. Moreover, the Pareto-optimal solutions can be viewed as the features of the decision space in solving MaOPs, and if we obtain the subspace of the PS, subsequent operations can be constrained in this subspace to reduce the cost of exploration. Basically, this subspace is obtained by reducing the dimension on the Pareto-optimal solutions. Ideally, only the diversity is concerned if the exact subspace of the PS is obtained, which is not true in implementation. To this end, extra components are incorporated in the proposed algorithm for improving the convergence, such as the reference vector updating, and non-dominated selection for building model, diversity repairing, generating offspring, environmental selection, and final selection.

\begin{figure}
\begin{center}
\subfloat[]{\includegraphics[width=1.5in]{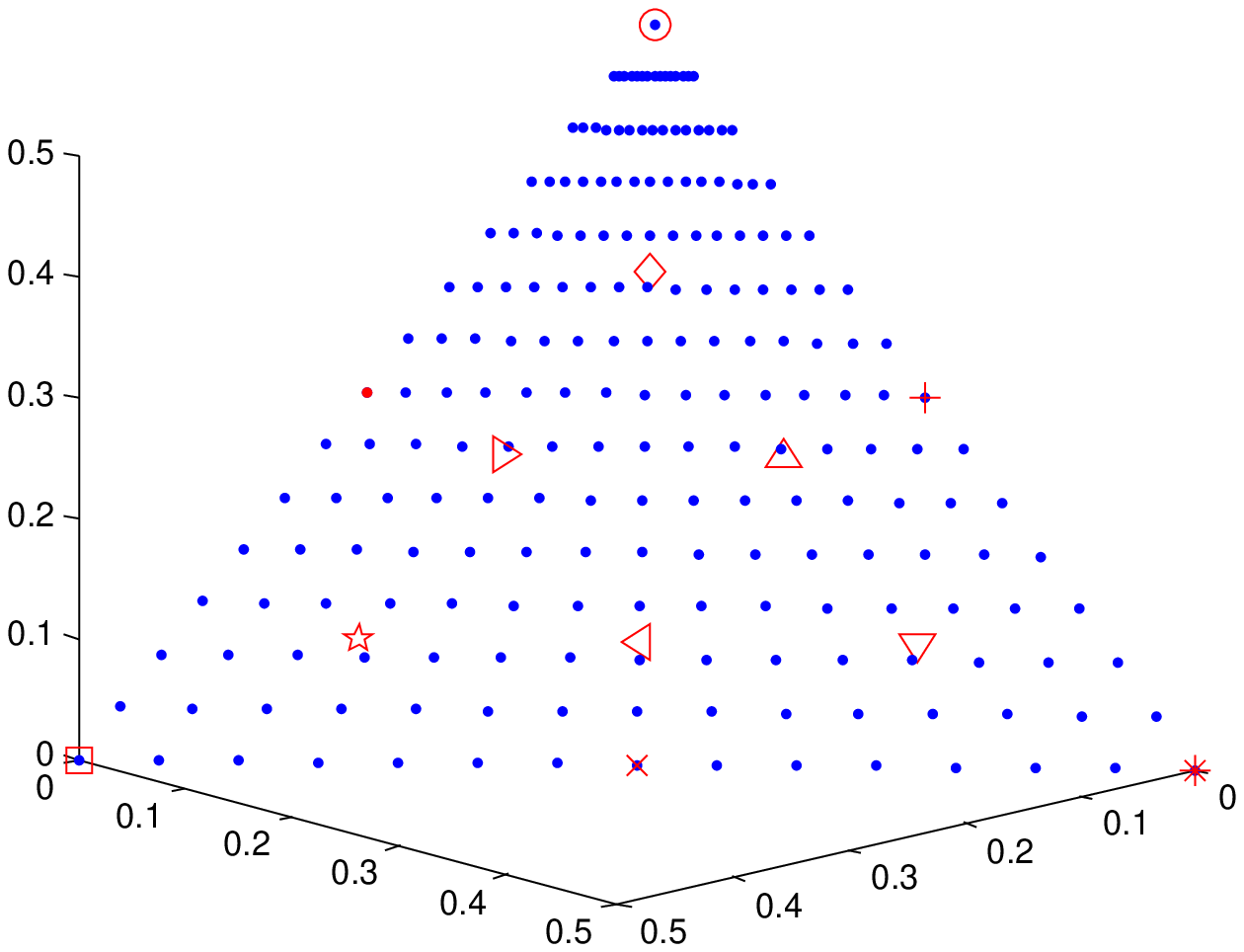}%
\label{fig_alg_1}}
\hfil
\subfloat[]{\includegraphics[width=1.5in]{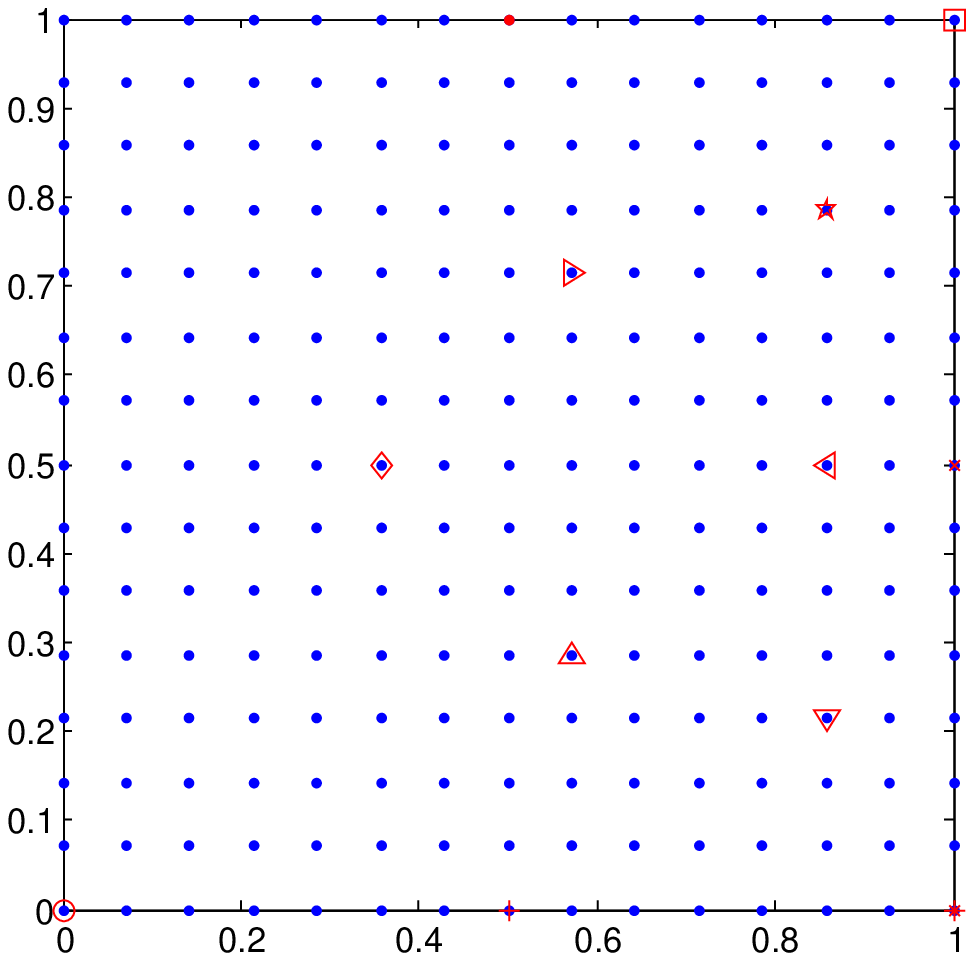}%
\label{fig_alg_2}}
\hfil
\subfloat[]{\includegraphics[width=1.5in]{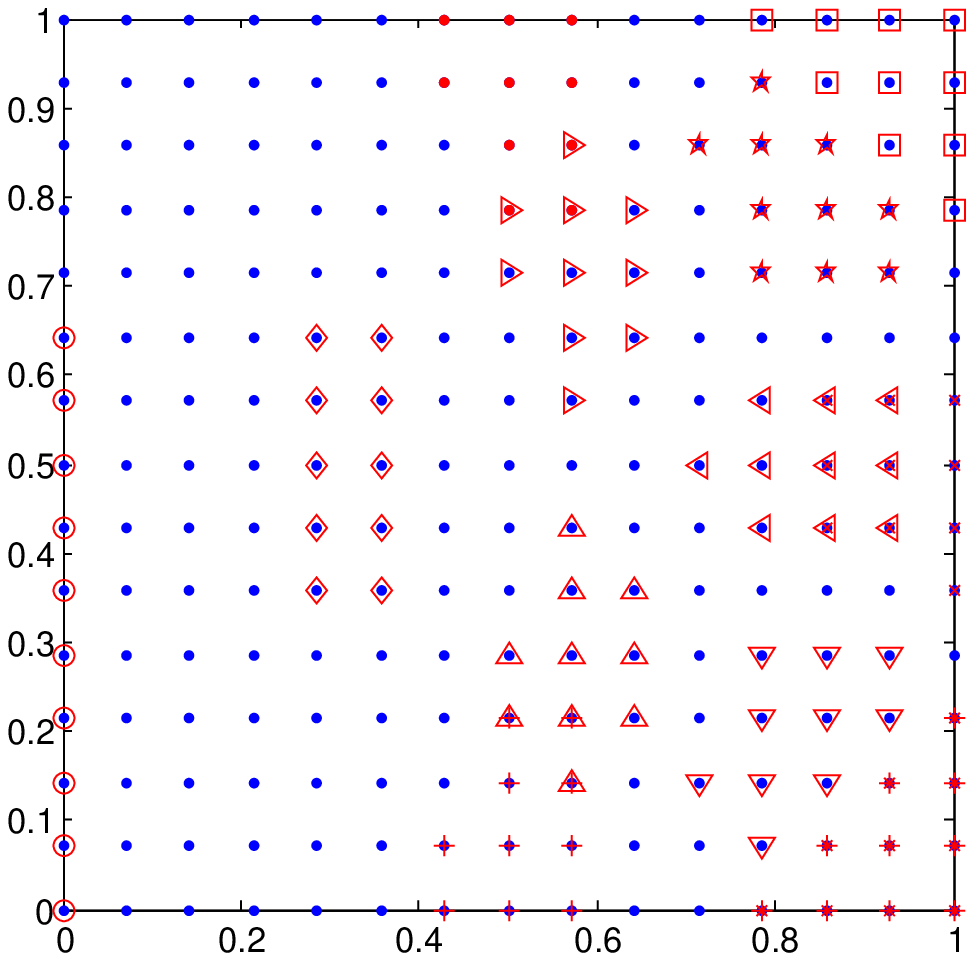}%
\label{fig_alg_3}}
\hfil
\subfloat[]{\includegraphics[width=1.5in]{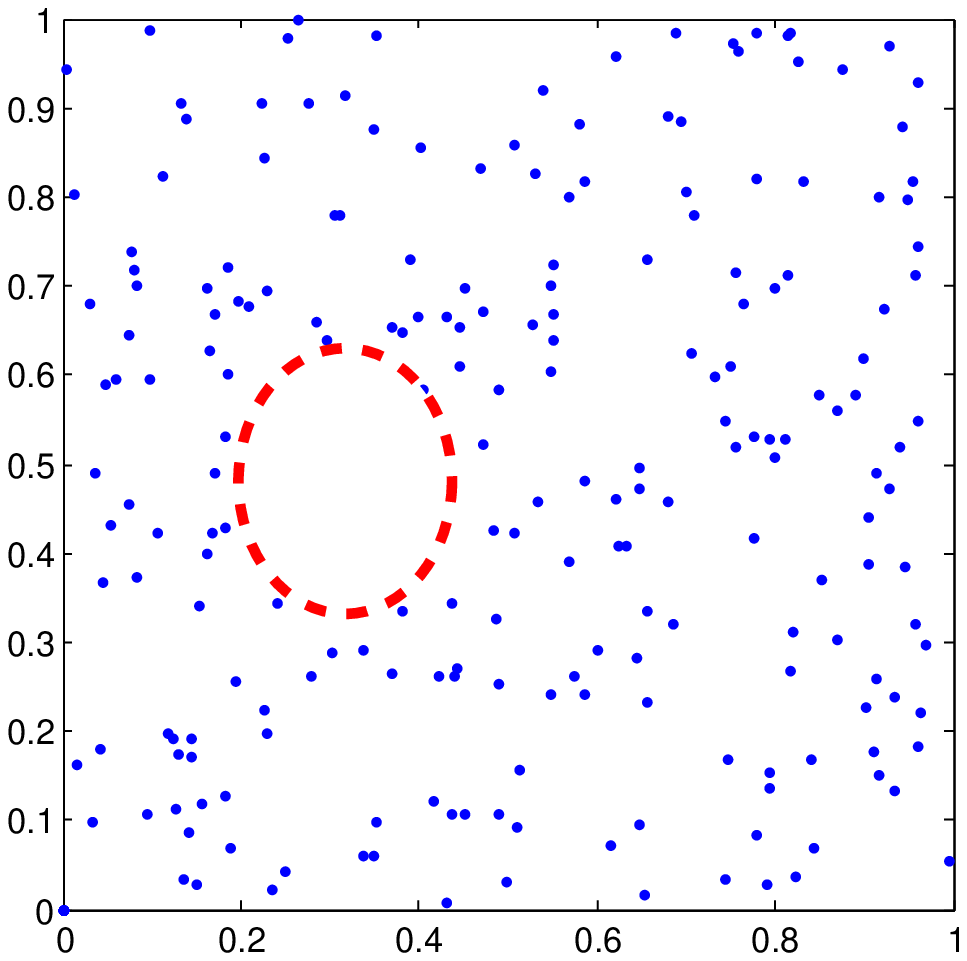}%
\label{fig_alg_4}}
\hfil
\subfloat[]{\includegraphics[width=1.5in]{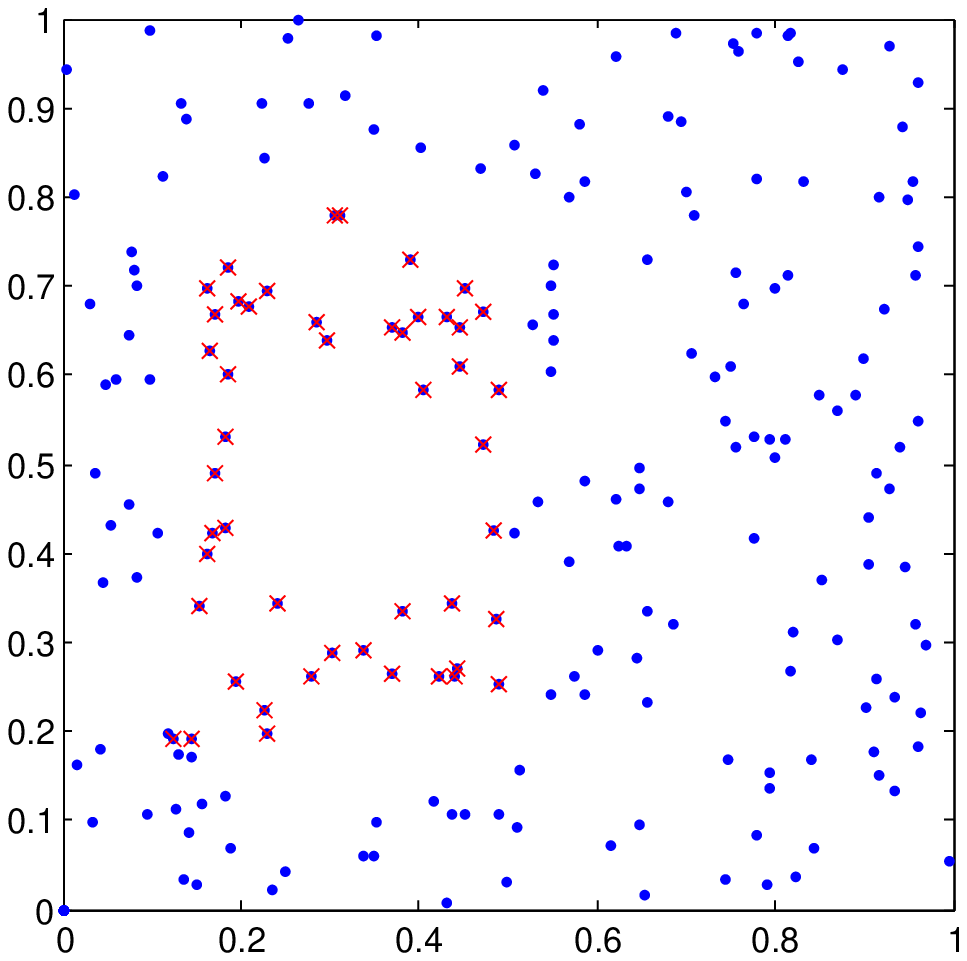}%
\label{fig_alg_5}}
\hfil
\subfloat[]{\includegraphics[width=1.5in]{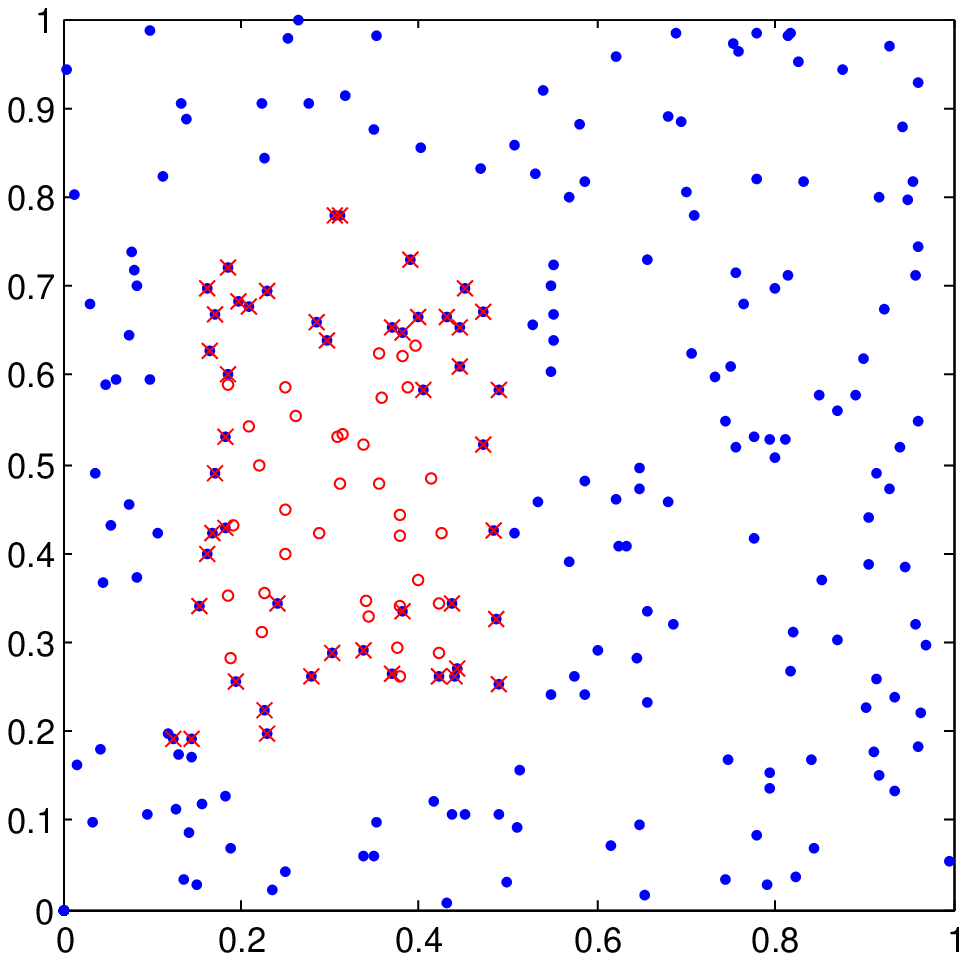}%
\label{fig_alg_6}}
\caption{A schematic diagram of the diversity repairing mechanism. In Figs.~\ref{fig_alg_2})-\ref{fig_alg_3}), the blue markers denote the uniformly distributed solutions in the decision variable space with the reduced dimension, and their corresponding objectives are plotted with the same color in Fig.~\ref{fig_alg_1}). A set of uniformly distributed reference vectors is plotted in Fig.~\ref{fig_alg_1}) in red color and their corresponding solutions and neighbors are plotted with the same shape in Figs.~\ref{fig_alg_2}) and \ref{fig_alg_3}), respectively. The blue markers in Figs.~\ref{fig_alg_4})-\ref{fig_alg_6}) denote all the solutions in the reduced dimension decision variable space in one generation. Especially, the area constrained by the ellipse circle denotes there is no solution for assigning to the corresponding reference vector. The cross markers in Fig.~\ref{fig_alg_5}) are the solutions selected for generating new solutions which are plotted in Fig.~\ref{fig_alg_6}) with solid markers to repair the diversity of the corresponding reference vector.}
\label{fig_diversity_repairing}
\end{center}
\end{figure}

To intuitively understand the diversity repairing mechanism in the proposed algorithm, an example of $3$-objective DTLZ1 problem~\cite{deb2005scalable} is illustrated in Fig.~\ref{fig_diversity_repairing}. To be specific, twelve different markers in red are shown in Fig.~\ref{fig_alg_1}) identifying a set of uniformly distributed reference vectors in the objective space. The corresponding solutions (given the same markers) in the decision space with the reduced dimensions are shown in Fig.~\ref{fig_alg_2}) which are not necessarily uniformly distributed. For each given solution in the decision space, nine different neighbors are chosen and displayed in the same marker as shown in Fig.~\ref{fig_alg_3}). However, all the reference vectors are not necessarily assigned by the solutions in one generation due to the heuristic nature of evolution. This can be seen from Fig.~\ref{fig_alg_4}) in which blue markers denote all the solutions in this generation and the area constrained by the ellipse circle implies that there is no solution for assigning to the corresponding reference vector. To this end, diversity repairing mechanism is activated by using the neighbor solutions. The ``cross'' markers in Fig.~\ref{fig_alg_5}) are the solutions selected for generating new solutions. These solutions are then used for diversity repairing and generate solutions shown in ``hollow circle'' markers in Fig.~\ref{fig_alg_6}). Intuitively, diversity repairing is employed for improving the diversity by assigning the corresponding solution. In fact, the convergence is strengthened in the phase of diversity repairing by generating new solutions from which dominated solution is assigned to the corresponding reference vector which is short of diversity previously. In summary, both convergence and diversity are promoted by diversity repairing mechanism.

\begin{figure}
\begin{center}
\subfloat[]{\includegraphics[width=0.6\columnwidth]{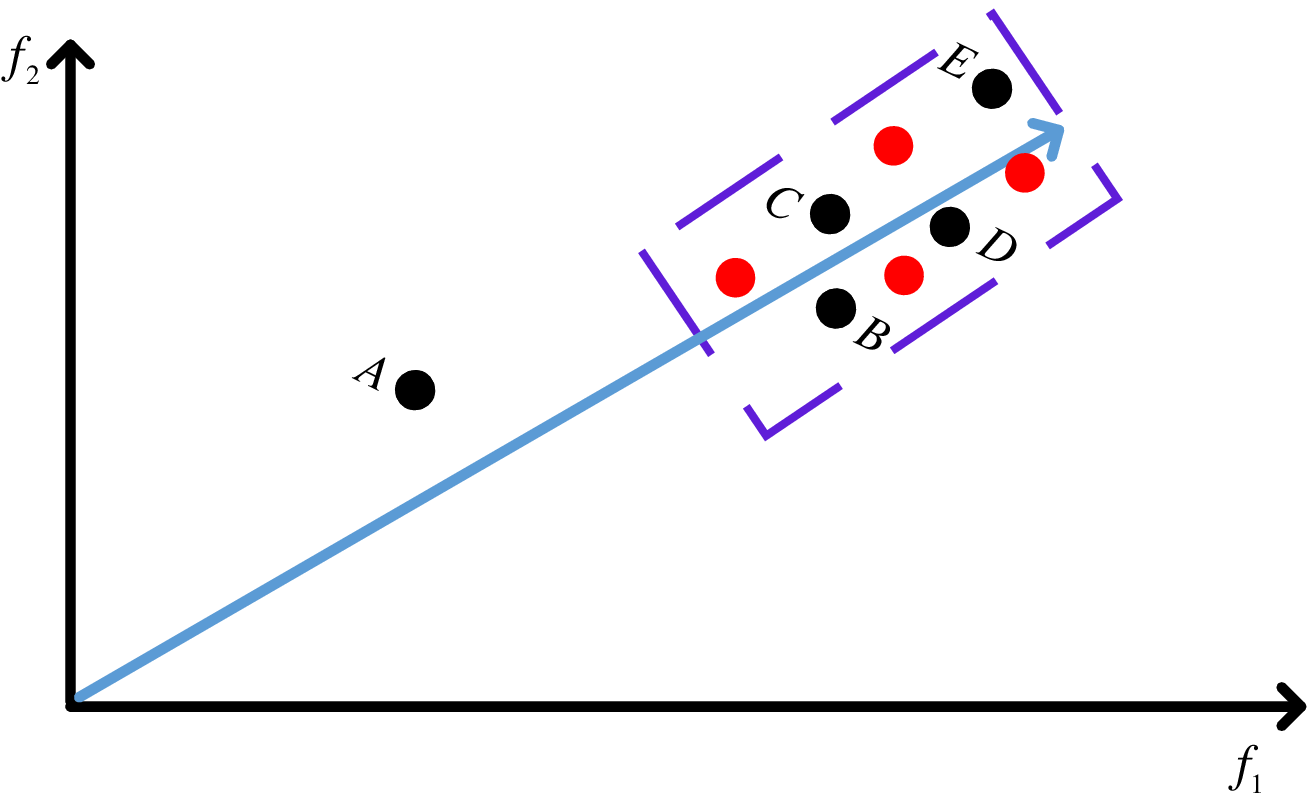}%
\label{fig_neighbour_assigment_1}}
\hfil
\subfloat[]{\includegraphics[width=0.6\columnwidth]{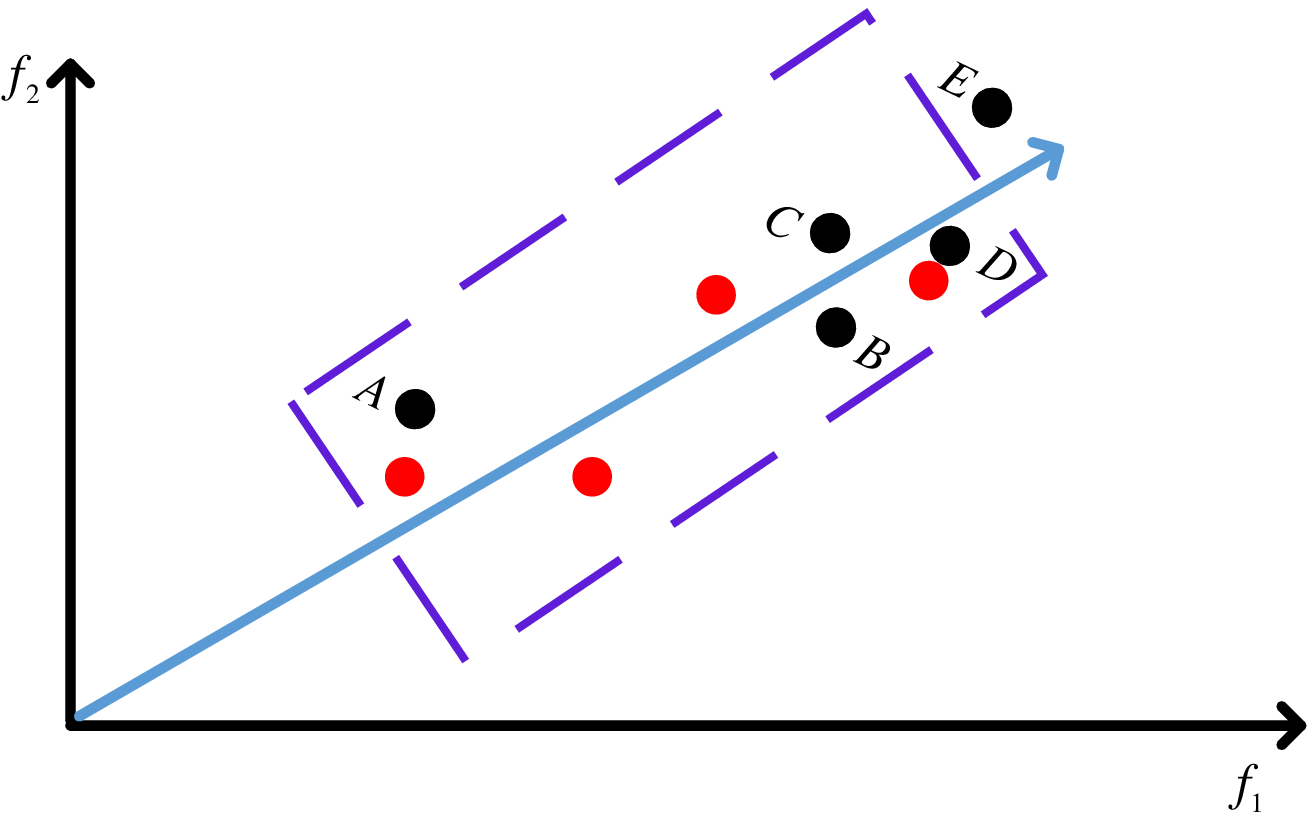}%
\label{fig_neighbour_assigment_2}}

\caption{An example to highlight the quality of generated solutions that the non-dominated solution $A$ is included in the neighbor solutions (see Fig.~\ref{fig_neighbour_assigment_2}) or not (see Fig.~\ref{fig_neighbour_assigment_1}). Especially, the blue line denotes the reference vector, $A, B, C, D$ and $E$ denote the current population, rectangle area denotes the built model based on the selected neighbor solutions, and red circles are the generated solutions.}
\label{fig_neighbour_assigment}
\end{center}
\end{figure}

Normally, the neighbor solutions for one particular reference vector consist of one non-dominated solution and other nearest solutions to this reference vector from the current population. With this neighbor solution assignment, it is hopeful that solutions with convergence and diversity are sampled from the built model. Moreover, Fig.~\ref{fig_neighbour_assigment} highlights our motivation on this design. Especially, the blue line denotes the reference vector, black circles, $A, B, C, D$ and $E$ denotes the current population, rectangle area denotes the built model, and red circles denoted the sampled offspring. Fig.~\ref{fig_neighbour_assigment_1} plots the solutions which have the nearest perpendicular distances to the reference vectors, while Fig.~\ref{fig_neighbour_assigment_2} depicts the non-dominated solution included into the neighbor solutions. It is obviously that solutions with good diversity and convergence are generated in Fig.~\ref{fig_neighbour_assigment_2}.

\section{Experiments}
\label{section_4}
To demonstrate the quality of the proposed algorithm, a series of experiments are well-designed and performed on $8$ test problems, which are from two benchmark test suits, DTLZ~\cite{deb2005scalable} and DTLZ$^{-1}$~\cite{ishibuchiperformance}, with $3$-, $5$-, $8$-, $10$-, and $15$-objective. Since the proposed MaOEDA-IR is an EDA-based algorithm for solving MaOPs, state-of-the-art algorithms covering two categories 1) traditional MaOEAs (NSGA-III~\cite{deb2014evolutionary}, MOEA/D~\cite{zhang2007moea}, GrEA~\cite{yang2013grid}, and HypE~\cite{bader2011hype}) and 2) EDA-based evolutionary algorithm (MBN-EDA~\cite{karshenas2014multiobjective}, and RM-MEDA~\cite{zhang2008rm}) are considered as the peer competitors to compare the performance against the proposed MaOEDA-IR.

In the following subsections, the selected benchmark test problems are introduced first. Then, the performance indicators chosen to measure the results generated by these compared algorithms are documented. Next, the parameter settings utilized by compared algorithms are declared. Finally, experiments on compared algorithms are performed and their results measured by the selected performance indicators are analyzed. In addition, empirical experiments on investigating the diversity repairing, dimension reduction, and neighbor size are performed to highlight their superiority and promote efficacy in addressing real-word problems.

\subsection{Benchmark Test Problems}
\label{sec_benchmark_test_problems}
DTLZ1-DTLZ4 problems which are from the scalable benchmark test suit DTLZ are considered as the test instances in these experiments. Specifically, each $M$-objective test problem is with $n=M+k-1$ decision variables where $k$ is specified as $5$ for DTLZ1 and $10$ for DTLZ2-DTLZ4. Furthermore, the Pareto-optimal solutions of DTLZ1-DTLZ4 in the normalized $M$-dimensional objective space have the form formulated by Equation~\ref{equ_pareto_optimal_solution_dtlz1-4}.
\begin{equation}
  \label{equ_pareto_optimal_solution_dtlz1-4}
  \sum_{i=1}^M f_i(x)^p = 1
\end{equation}
where $p=1$ for DTLZ1 and $p=2$ for DTLZ2-DTLZ4.
Because the employed reference vectors $r_{0,i}=[r_{0,i}^1,\cdots, r_{0,i}^M]$ generated by the systematical Das and Dennis's method in the proposed algorithm are with the form $\sum_{j=1}^Mr_{0,i}^j=1$ that is similar to Equation~\ref{equ_pareto_optimal_solution_dtlz1-4}, DTLZ test problems are considered less challengeable. To this end, the DTLZ1$^{-1}$-DTLZ4$^{-1}$ problems from the DTLZ$^{-1}$ test suit, which is a variant of the DTLZ by multiplying the negative sign to each test problem in DTLZ, are included into the considered benchmark test problems for their more complicated PF shapes which especially challenges algorithms based on reference vectors.
\subsection{Performance Metrics}
\label{sec_performance_metrics}
Two widely used performance metrics, Inverted Generational Distance (IGD)~\cite{bosman2003balance} and Hypervolume (HV)~\cite{zitzler1999multiobjective} which can simultaneously quantify the performance in convergence and diversity of the algorithms, are adopted in these experiments. The results generated by these compared algorithms are normalized to $[0,1]$ priori to employing the performance indicators, which is in the same manner~\cite{yuan2016new}. In addition, $100,000$ reference points are uniformly sampled from Equation~\ref{equ_pareto_optimal_solution_dtlz1-4} for the calculation of IGD, and $[1.1,\cdots,1.1]$ is specified as the reference point for the calculation of HV. Furthermore, Monte Carlo simulation~\cite{bader2011hype} is applied for the calculation of HV when $M\geq 10$, otherwise the exact approach proposed in~\cite{while2012fast} is utilized due to the computation cost dramatically increasing as the number of objectives grows.

\subsection{Parameter Settings}
\label{sec_parameter_settings}
In this subsection, the parameter settings are presented. First, the general settings for most compared algorithms are listed. Thereafter, special settings for partial algorithms are specified.

\subsubsection{Crossover and Mutation} SBX~\cite{deb1995simulated} and polynomial mutations~\cite{deb2001multi} are employed as the crossover operator and mutation operator, respectively. Furthermore, the probabilities for SBX and polynomial mutation, and the crossover distribution index are set to be $1.0$, $1/n$ ($n$ is the number of decision variables), and $20$, respectively. In addition, the distribution index of NSGA-III is set to be $30$ according to the suggestions in~\cite{deb2014evolutionary} while others are set to be $20$.
\subsubsection{Population Size} The population size can be set arbitrarily for executing experiments. However, reference vectors assisted algorithms, such as NSGA-III, require the same number of the population size to that of reference vectors, other peer algorithms adopt the same population size for a fair comparison. Furthermore, only boundary reference vectors are generated when the division numbers is less than $M$ in the phase of sampling of reference vectors. To this end, the two-layer approach~\cite{deb2014evolutionary} is employed for generating the reference vectors. In addition, the implementation of GrEA and NSGA-III require the population size to be a multiple of $4$. In summary, the settings for reference vector and population size are listed in Table~\ref{table_population_size}.
\begin{table}[ht]
\caption{Settings for reference vectors and population size.}
\label{table_population_size}
\begin{center}
\begin{tabular}{p{0.05\columnwidth}<{\centering}|p{0.06\columnwidth}<{\centering}|p{0.06\columnwidth}<{\centering}|p{0.2\columnwidth}<{\centering}|p{0.35\columnwidth}<{\centering}}
\hline
\multirow{2}{*}{$M$}& \multicolumn{2}{c|}{\# of division} & \# of reference  &population size of\\
\cline{2-3}
& $H_1$ & $H_2$ & vectors&  GrEA and NSGA-III\\
\hline
3 & 14 & - & 120 & 120 \\
5 & 5 & - & 126 & 128 \\
8 & 3 & 2 & 156 & 156\\
10 & 2 & 2 & 110 & 112 \\
15 & 2 & 1 & 135 &  136 \\
\hline
\end{tabular}
\end{center}
\end{table}
\subsubsection{Special Settings}
The grid sizes of GrEA varying in $\{6,7,8,9,10,11,12\}$ are tested individually and the best scores selected based on the corresponding performance indicators are picked up for comparisons. Because RM-MEDA is originally designed for MOEAs, the default configurations are not suitable for solving MaOPs in this experiments. Consequently, the parameter setting in RM-MEDA is slightly modified for maximizing its performance to deal with MaOPs. Specifically, the number of clusters in local PCA varies in $\{10, 20, 30, 50\}$; the maximum iterations of local PCA with $\{50,100\}$ are tested individually with the maximum generations at $500$ for selecting the best mean value indicated by performance metrics, while the parameters settings in MBN-EDA are set based on the developers' suggestion in~\cite{karshenas2014multiobjective}. In addition, both the thresholds for dimension reduction and model building are set to be $0.96$ according to the convention of the community. Furthermore the neighbor size is specified as $25$ for the investigation in Subsection~\ref{sec_investigation_neighbor_size}, and the enlargement factor is set to be $0.5$. In the settings of the proposed algorithm, the generation number of PCSEA for obtaining training data is specified as $50$, and the population size is set to be $100$.

The proposed MaOEDA-IR is based on the training data obtained by PCSEA. For a fair comparison, the termination criterion of MaOEDA-IR should be set to the total function evaluation number of the peer competitors minus that of PCSEA. Table~\ref{tab_function_evaluation}\footnote{These settings apply only to the experimental results in Subsection~\ref{sub_results}. For other experiments, the terminated criteria for MaOEDA-IR and PCSEA are specified as $200$ generations, and the population size for PCSEA is set to be $100$ for $M\leq 10$ and $200$ for $M >10$.} shows these particular settings for each considered number of objective.
\begin{table}[ht]
\caption{The settings for the maximal function evaluation numbers.}
\label{tab_function_evaluation}
\begin{center}
\begin{tabular}{c|c|c|c}
\hline
\multirow{2}{*}{$M$}& total evaluation & evaluation numbers &evaluation numbers\\
& numbers & for PCSEA & for MaOEDA-IR\\
\hline
3	& 7.2E+4	& 1.5E+4	& 5.7E+4 \\

5 & 1.3E+5 & 2.5E+4& 1.0E+5\\

8	& 2.5E+5	& 4.0E+4 & 2.1E+5 \\

10 & 2.2E+5 & 5.0E+4 & 1.7E+5 \\

15 & 4.1E+5 & 7.5E+4 & 3.3E+5\\
\hline
\end{tabular}
\end{center}
\end{table}

\subsection{Performance on DTLZ and DTLZ$^{-1}$}
\label{sub_results}
The HV results of the proposed MaOEDA-IR against its peer competitors (NSGA-III, MOEA/D, GrEA, HypE, MBN-EDA, and RM-MEDA) on DTLZ1-DTLZ4 and DTLZ1$^-$-DTLZ4$^-$ test problems with $3$-, $5$-, $8$-, $10$-, and $15$-objective are presented in Table~\ref{hv_results_on_dtlz1_7}. Furthermore, each compared algorithm is independently performed $30$ runs and the best median HV results are highlighted in bold face. Moreover, the Mann-Whitney-Wilcoxon rank-sum test~\cite{steel1997principles} with a $5\%$ significance level is used to conduct the HV results due to the heuristic nature of peer algorithms, and the symbols ``+,'' ``=,'' and ``-'' denote whether the HV results of the proposed MaOEDA-IR are statistically better than, equal to, or worse than those of the corresponding peer competitors. In addition, the last row in Table~\ref{hv_results_on_dtlz1_7} summarizes how many times the proposed MaOEDA-IR is better than, equal to, or worse than its respective peer competitor.

The results in Table~\ref{hv_results_on_dtlz1_7} indicate that the proposed MaOEDA-IR obtains the best performances on the DTLZ4, DTLZ1$^-$, and DTLZ3$^-$ test problems with all considered objective numbers except for $5$-objective DTLZ1$^-$ and $8$-objective DTLZ3$^- $ which are worse than GrEA. Moreover, MaOEDA-IR is superior to others over DTLZ3 with $8$- and $10$-objective while is inferior to GrEA with $3$-objective, RM-MEDA with $5$-objective, and NSGA-III with $15$-objective. Although the performance of MaOEDA-IR in DTLZ2 and DTLZ4$^-$ are worse than those of NSGA-III and GrEA on the $3$-, and $5$-objective, respectively, the proposed MaOEDA-IR performs better than others on the $10$-objective. In addition, the proposed MaOEDA-IR outperforms peer competitors on DTLZ1 and DTLZ2$^-$ test problems with $8$-, and $15$-objective. In summary, the HV results of the proposed MaOEDA-IR against selected compared algorithms over eight test problems with $3$-, $5$-, $8$-, $10$-, and $15$-objective indicate that MaOEDA-IR has a comparable performance by winning $181$ test scores out of $240$ comparisons, and performing equally well to $10$ comparisons.

\begin{table*}[ht]
\caption{HV results of MaOEDA-IR against NSGA-III, MOEA/D, GrEA, HypE, MBN-EDA, and RM-MEDA over the DTLZ1, DTLZ2, DTLZ3, DTLZ4, DTLZ1$^-$, DTLZ2$^-$, DTLZ3$^-$, and DTLZ4$^-$ test problems with $3$-, $5$-, $8$-, $10$-, and $15$-objective. Each compared algorithm is independently performed $30$ runs, and the best median HV results are highlighted in bold face. The symbols ``+,'' ``=,'' and ``-'' denote whether the HV results of the proposed MaOEDA-IR are statistically better than, equal to, or worse than that of the corresponding peer competitors with a significant level $5\%$, respectively.}
\label{hv_results_on_dtlz1_7}
\begin{center}
\begin{tabular}{c|c|c|c|c|c|c|c|c}
\hline
Problem&$M$&MaOEDA-IR&NSGA-III&MOEA/D&GrEA&HypE&MBN-EDA&RM-MEDA\\
\hline
\multirow{5}{*}{DTLZ1}
&3&0.838(2.1E-2)&0.915(5.6E-2)(-)&0.816(1.6E-3)(+)&0.925(8.2E-2)(-)&0.906(6.3E-2)(-)&\textbf{0.999(9.8E-4)(-)}&0.835(2.5E-2)(=)\\
\cline{2-9}
&5&0.959(3.3E-3)&0.986(8.9E-3)(-)&0.936(2.6E-3)(+)&\textbf{0.989(8.0E-3)(-)}&0.973(3.0E-2)(-)&0.970(6.9E-3)(-)&0.980(9.5E-5)(-)\\
\cline{2-9}
&8&\textbf{0.997(6.1E-4)}&0.996(3.1E-3)(=)&0.202(1.1E-1)(+)&0.988(5.0E-2)(+)&0.982(1.8E-2)(+)&0.984(3.5E-3)(+)&0.997(2.4E-3)(=)\\
\cline{2-9}
&10&0.991(4.2E-4)&0.990(4.0E-3)(+)&0.047(6.7E-2)(+)&0.971(2.4E-2)(+)&0.937(5.1E-2)(+)&0.984(4.1E-3)(+)&\textbf{0.992(5.2E-3)(-)}\\
\cline{2-9}
&15&\textbf{0.997(2.2E-4)}&0.994(6.1E-3)(+)&0.153(1.5E-1)(+)&0.968(3.0E-2)(+)&0.930(4.3E-2)(+)&0.992(2.3E-3)(+)&0.993(4.2E-3)(+)\\
\hline
\multirow{5}{*}{DTLZ2}
&3&0.533(1.2E-3)&\textbf{0.645(4.7E-2)(-)}&0.540(2.1E-3)(-)&0.543(1.9E-3)(-)&0.361(2.9E-2)(+)&0.563(4.5E-2)(-)&0.566(1.6E-3)(-)\\
\cline{2-9}
&5&0.780(5.5E-3)&\textbf{0.956(1.1E-2)(-)}&0.710(1.8E-3)(+)&0.795(1.2E-3)(-)&0.494(6.4E-2)(+)&0.739(4.0E-2)(+)&0.795(1.0E-2)(-)\\
\cline{2-9}
&8&\textbf{0.925(6.7E-4)}&0.924(4.3E-3)(+)&0.925(8.7E-3)(=)&0.904(2.8E-3)(+)&0.543(8.6E-2)(+)&0.842(3.9E-2)(+)&0.020(1.1E-2)(+)\\
\cline{2-9}
&10&\textbf{0.932(9.3E-3)}&0.931(1.0E-2)(+)&0.002(4.1E-3)(+)&0.922(3.7E-3)(+)&0.443(7.2E-2)(+)&0.879(3.1E-2)(+)&0.880(1.8E-2)(+)\\
\cline{2-9}
&15&0.979(4.2E-3)&\textbf{0.982(4.4E-4)(-)}&0.066(1.1E-1)(+)&0.888(1.8E-2)(+)&0.404(7.2E-2)(+)&0.922(1.9E-2)(+)&0.876(1.6E-2)(+)\\
\hline
\multirow{5}{*}{DTLZ3}
&3&0.513(2.7E-3)&0.986(3.0E-2)(-)&0.574(7.3E-2)(-)&\textbf{0.979(2.2E-4)(-)}&0.816(1.7E-1)(-)&0.680(4.6E-2)(-)&0.977(7.1E-3)(-)\\
\cline{2-9}
&5&0.989(3.6E-4)&0.987(1.8E-2)(+)&0.704(2.8E-2)(+)&0.794(1.7E-3)(+)&0.995(3.7E-3)(-)&0.862(3.1E-2)(+)&\textbf{0.996(1.4E-3)(-)}\\
\cline{2-9}
&8&\textbf{0.997(2.6E-5)}&0.994(6.5E-3)(+)&0.313(1.9E-1)(+)&0.923(5.6E-4)(+)&0.995(6.2E-3)(+)&0.921(1.5E-2)(+)&0.949(8.9E-3)(+)\\
\cline{2-9}
&10&\textbf{0.989(2.0E-5)}&0.984(1.4E-2)(+)&0.096(1.4E-1)(+)&0.943(8.2E-4)(+)&0.988(1.0E-2)(+)&0.929(1.4E-2)(+)&0.898(1.9E-2)(+)\\
\cline{2-9}
&15&0.975(4.9E-4)&\textbf{0.997(3.0E-3)(-)}&0.079(9.3E-2)(+)&0.902(2.2E-2)(+)&0.985(1.2E-2)(-)&0.963(4.1E-3)(+)&0.887(1.6E-2)(+)\\
\hline
\multirow{5}{*}{DTLZ4}
&3&\textbf{0.733(2.9E-4)}&0.521(9.2E-2)(+)&0.445(1.0E-1)(+)&0.476(1.7E-1)(+)&0.471(1.0E-1)(+)&0.565(2.1E-3)(+)&0.538(3.2E-3)(+)\\
\cline{2-9}
&5&\textbf{0.920(6.9E-3)}&0.798(9.0E-3)(+)&0.614(5.9E-2)(+)&0.785(3.0E-2)(+)&0.585(6.6E-2)(+)&0.795(1.1E-3)(+)&0.726(2.8E-2)(+)\\
\cline{2-9}
&8&\textbf{0.938(5.2E-3)}&0.923(2.3E-3)(+)&0.065(5.7E-2)(+)&0.916(2.1E-3)(+)&0.521(9.2E-2)(+)&0.928(9.5E-4)(+)&0.859(1.4E-2)(+)\\
\cline{2-9}
&10&\textbf{0.953(6.2E-5)}&0.943(2.3E-3)(+)&0.011(2.1E-2)(+)&0.936(1.3E-3)(+)&0.449(9.4E-2)(+)&0.949(9.7E-4)(+)&0.897(2.1E-2)(+)\\
\cline{2-9}
&15&\textbf{0.997(3.2E-5)}&0.991(5.5E-4)(+)&0.009(1.9E-2)(+)&0.936(9.6E-3)(+)&0.546(9.1E-2)(+)&0.991(2.3E-4)(+)&0.985(9.4E-3)(+)\\
\hline

\multirow{5}{*}{DTLZ1$^-$}
&3&\textbf{0.289(3.1E-3)}&0.217(3.2E-3)(+)&0.208(3.2E-3)(+)&0.227(1.3E-3)(+)&0.121(1.0E-2)(+)&0.118(1.6E-2)(+)&0.255(1.6E-2)(+)\\
\cline{2-9}
&5&0.010(2.7E-3)&0.019(3.8E-3)(-)&0.007(4.1E-4)(+)&\textbf{0.199(9.0E-3)(-)}&0.001(1.0E-4)(+)&0.019(3.7E-3)(-)&0.011(1.0E-3)(=)\\
\cline{2-9}
&8&\textbf{0.112(2.9E-2)}&0.001(5.2E-4)(+)&0.000(1.2E-5)(+)&0.000(1.4E-5)(+)&0.000(3.1E-6)(+)&0.001(0.0E+0)(+)&0.000(2.6E-5)(+)\\
\cline{2-9}
&10&\textbf{0.099(3.6E-4)}&0.000(1.7E-4)(+)&0.000(0.0E+0)(+)&0.000(0.0E+0)(+)&0.000(0.0E+0)(+)&0.000(0.0E+0)(+)&0.000(6.6E-6)(+)\\
\cline{2-9}
&15&\textbf{0.102(6.6E-3)}&0.000(0.0E+0)(+)&0.000(0.0E+0)(+)&0.000(0.0E+0)(+)&0.000(0.0E+0)(+)&0.000(0.0E+0)(+)&0.000(0.0E+0)(+)\\
\hline
\multirow{5}{*}{DTLZ2$^-$}
&3&0.533(7.2E-2)&0.541(1.0E-2)(-)&0.527(1.8E-3)(+)&\textbf{0.587(2.3E-2)(-)}&0.465(2.5E-2)(+)&0.340(2.3E-2)(+)&0.531(6.1E-3)(+)\\
\cline{2-9}
&5&0.067(3.3E-5)&0.140(2.0E-2)(-)&0.078(2.0E-3)(-)&\textbf{0.258(3.1E-2)(-)}&0.008(1.3E-3)(+)&0.144(2.2E-2)(-)&0.067(9.7E-3)(=)\\
\cline{2-9}
&8&\textbf{0.102(5.2E-3)}&0.012(4.5E-3)(+)&0.000(4.5E-5)(+）&0.001(8.4E-5)(+)&0.000(1.6E-5)(+)&0.017(3.4E-3)(+)&0.001(2.1E-4)(+)\\
\cline{2-9}
&10&\textbf{0.033(3.6E-2)}&0.004(2.4E-3)(+)&0.000(0.0E+0)(+)&0.000(1.5E-5)(+)&0.000(0.0E+0)(+)&0.002(5.7E-5)(+)&0.000(2.4E-5)(+)\\
\cline{2-9}
&15&\textbf{0.003(3.2E-3)}&0.000(4.2E-5)(+)&0.000(0.0E+0)(+)&0.000(0.0E+0)(+)&0.000(0.0E+0)(+)&0.000(0.0E+0)(+)&0.000(0.0E+0)(+)\\
\hline
\multirow{5}{*}{DTLZ3$^-$}
&3&\textbf{0.553(1.9E-2)}&0.539(1.1E-2)(+)&0.535(1.7E-2)(+)&0.540(1.7E-3)(+)&0.483(1.5E-2)(+)&0.242(2.9E-2)(+)&0.547(2.4E-2)(+)\\
\cline{2-9}
&5&\textbf{0.230(3.3E-2)}&0.131(1.9E-2)(+)&0.119(1.1E-2)(+)&0.071(9.0E-4)(+)&0.074(3.6E-3)(+)&0.105(1.2E-2)(+)&0.087(8.6E-3)(+)\\
\cline{2-9}
&8&0.001(3.2E-3)&0.012(3.0E-3)(-)&0.001(1.8E-4)(=)&\textbf{0.102(1.2E-2)(-)}&0.001(2.3E-4)(=)&0.014(4.0E-3)(-)&0.001(2.2E-4)(=)\\
\cline{2-9}
&10&\textbf{0.030(3.0E-3)}&0.003(2.0E-3)(+)&0.000(6.9E-6)(+)&0.000(1.4E-5)(+)&0.000(2.4E-5)(+)&0.003(1.1E-4)(+)&0.000(2.8E-5)(+)\\
\cline{2-9}
&15&\textbf{0.003(2.1E-5)}&0.000(1.6E-4)(+)&0.000(0.0E+0)(+)&0.000(0.0E+0)(+)&0.000(0.0E+0)(+)&0.000(0.0E+0)(+)&0.000(0.0E+0)(+)\\
\hline
\multirow{5}{*}{DTLZ4$^-$}
&3&0.531(2.8E-3)&0.533(6.4E-3)(-)&0.528(2.0E-3)(+)&\textbf{0.621(3.2E-2)(-)}&0.492(1.4E-2)(+)&0.170(1.2E-2)(+)&0.511(1.2E-2)(+)\\
\cline{2-9}
&5&0.060(3.9E-3)&0.076(1.5E-2)(-)&0.071(2.1E-3)(-)&\textbf{0.329(3.4E-2)(-)}&0.007(1.4E-3)(+)&0.005(1.1E-3)(+)&0.014(2.7E-3)(+)\\
\cline{2-9}
&8&0.001(6.3E-3)&0.002(1.0E-3)(-)&0.000(2.4E-5)(+)&\textbf{0.124(1.1E-2)(-)}&0.002(7.9E-6)(-)&0.000(0.0E+0)(+)&0.001(1.5E-5)(=)\\
\cline{2-9}
&10&\textbf{0.033(3.2E-3)}&0.000(2.9E-4)(+)&0.000(0.0E+0)(+)&0.000(3.1E-6)(+)&0.000(2.2E-6)(+)&0.000(0.0E+0)(+)&0.001(2.2E-6)(+)\\
\cline{2-9}
&15&\textbf{0.005(6.2E-2)}&0.003(3.8E-5)(+)&0.000(0.0E+0)(+)&0.000(0.0E+0)(+)&0.000(0.0E+0)(+)&0.000(0.0E+0)(+)&0.000(0.0E+0)(+)\\
\hline
\multicolumn{2}{c|}{+/=/-}&&25/1/14 & 34/2/4& 28/0/12& 33/1/6& 33/0/7& 28/6/6\\
\hline
\end{tabular}
\end{center}
\end{table*}

\subsection{Investigation on Neighbor Size}
\label{sec_investigation_neighbor_size}
\begin{figure}
  \centering
  \includegraphics[width=\columnwidth]{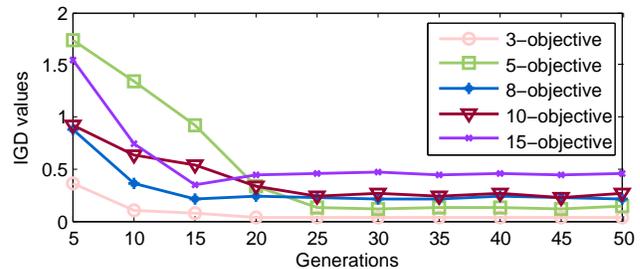}\\
  \caption{IGD values of the results generated by $3$-, $5$-, $8$-, $10$-, and $15$-objective DTLZ1 test problems with different neighbor sizes varying in $\{5,10,15,20,25,30,35,40,45,50\}$.}\label{fig_varying_t_size}
\end{figure}

To investigate how the neighbor size $T$ affecting the performance of the proposed MaOEDA-IR, a series of experiments is performed by varying $T$ in $[5,50]$ with an interval of $5$. Specifically, the results measured by IGD on DTLZ1 test problems with considered objective numbers are plotted in Fig.~\ref{fig_varying_t_size} in which it is clearly shown that appreciable changes have been taken place in $T$ with smaller numbers and gradually remain steady as $T$ increases. It is interpreted that the neighbor solutions with one particular reference vector for building the model are from other reference vectors when $T$ is with a smaller number, which causes the inaccuracy of the built model based on which new solutions are generated that led to deteriorating ues. Especially, most IGD values remain level when $T>25$ in Fig.~\ref{fig_varying_t_size} (it is actually applicable to other tested benchmark problems based on the investigations.), and the $T$ with larger size will increase the computational cost by introducing more initialized solutions. As a consequence, $T$ is specified as $25$ in our experiments.

\subsection{Investigation on Diversity Repairing Mechanism}
\label{sec_investigation_diversity_raparation}
\begin{figure}[htp]
\begin{center}
\subfloat[]{\includegraphics[width=0.8\columnwidth]{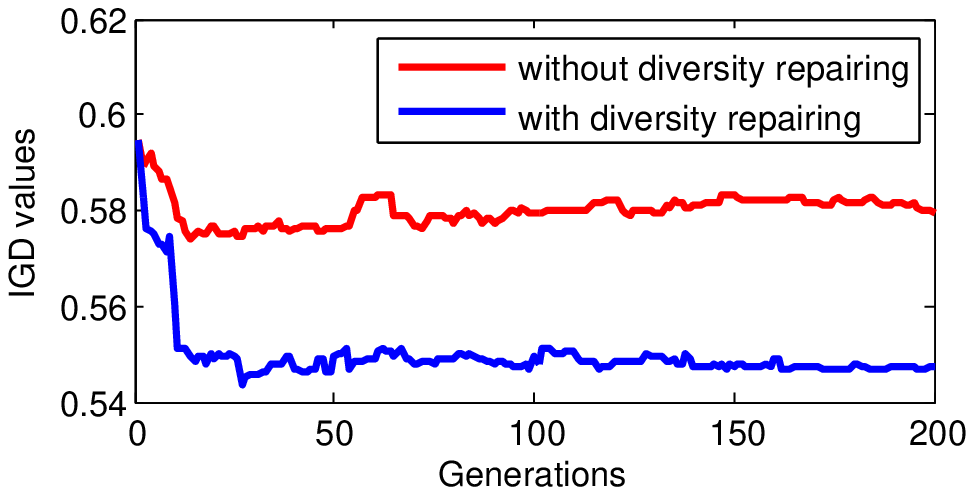}%
\label{fig_diversity_10}}
\hfil
\subfloat[]{\includegraphics[width=0.8\columnwidth]{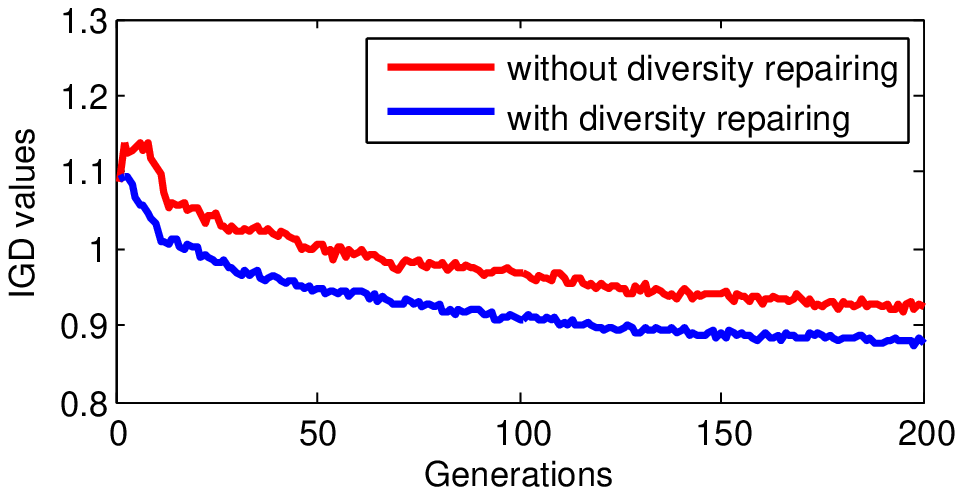}%
\label{fig_diversity_15}}

\caption{IGD values of the results generated by $10$-(Fig.~\ref{fig_diversity_10}) and $15$-objective (Fig.~\ref{fig_diversity_15}) DTLZ2 with and without diversity repairing over $200$ generations.}
\label{fig_diversity_comparison}
\end{center}
\end{figure}

\begin{figure}[htp]
\begin{center}
\subfloat[]{\includegraphics[width=0.8\columnwidth]{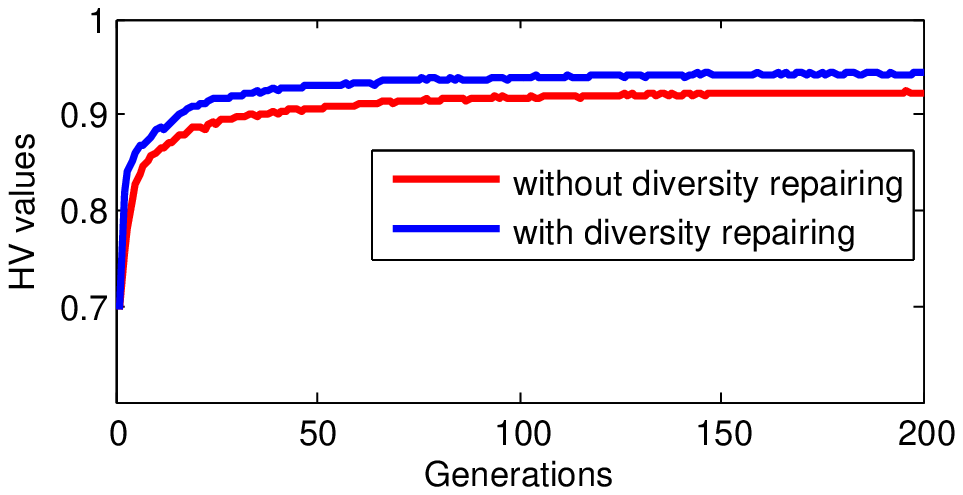}%
\label{fig_diversity_10_hv}}
\hfil
\subfloat[]{\includegraphics[width=0.8\columnwidth]{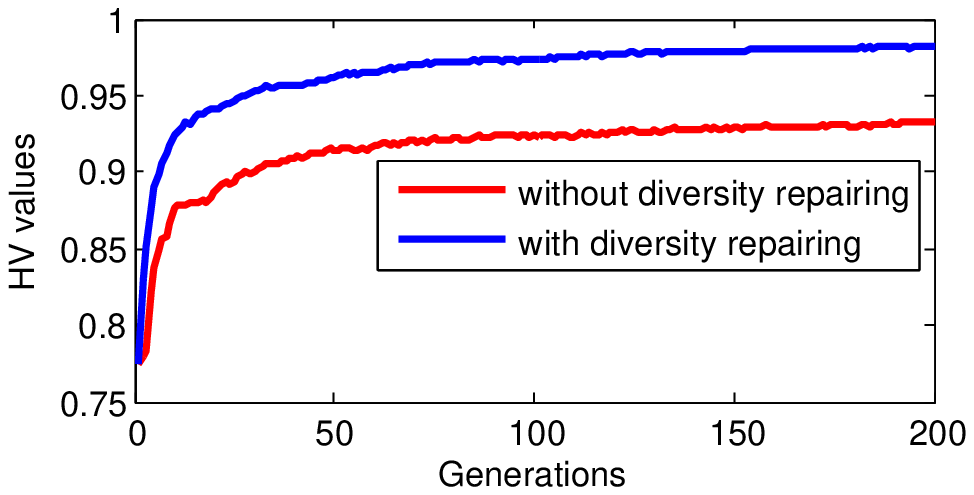}%
\label{fig_diversity_15_hv}}

\caption{HV values of the results generated by $10$-(Fig.~\ref{fig_diversity_10_hv}) and $15$-objective (Fig.~\ref{fig_diversity_15_hv}) DTLZ2 with and without diversity repairing over $200$ generations.}
\label{fig_diversity_comparison_hv}
\end{center}
\end{figure}

As discussed in Subsection~\ref{sec_discussion}, both the diversity and convergence have been improved with the diversity repairing mechanism. To this end, experimental comparisons on the test problems with and without the diversity repairing mechanism are performed. Specifically, the IGD values and the HV values of the evolution trajectory results generated by $10$- and $15$-objective DTLZ2 test problems over $200$ generations are illustrated in Figs.~\ref{fig_diversity_10}, ~\ref{fig_diversity_15},~\ref{fig_diversity_10_hv}, and~\ref{fig_diversity_15_hv},  respectively. In these figures, the red lines denote the results without the diversity repairing mechanism, while the blue lines refer to those with the diversity repairing. To be specific, both the IGD values of $10$-objective DTLZ2 with and without the diversity repairing mechanism sharply decrease during the first $20$ generations then gradually remain stable, while those of $15$-objective DTLZ2 smoothly decline throughout the entire evolution. For both the HV values of $10$-objective DTLZ2 with and without the diversity repairing mechanism, they grow substantially before the $40$-th generation then go up moderately as the evolution continues, while the ones resulted by the proposed algorithm without the diversity repairing mechanism stay lower than those with the diversity repairing mechanism during the entire evolution. In summary, both the best IGD results in Figs.~\ref{fig_diversity_10} and~\ref{fig_diversity_15} and the HV results in Figs.~\ref{fig_diversity_10_hv} and~\ref{fig_diversity_15_hv} demonstrate the promising performance of the proposed MaOEDA-IR when the diversity repairing mechanism is employed.

\begin{table}
\caption{HV results of MaOEDA-IR with and without the diversity repairing mechanism over the DTLZ1, DTLZ2, DTLZ3, DTLZ4, DTLZ1$^-$, DTLZ2$^-$, DTLZ3$^-$, and DTLZ4$^-$ test problems with $3$-, $5$-, $8$-, $10$-, and $15$-objective. Each compared algorithm is independently performed $30$ runs, and the best median HV results are highlighted in bold face. The symbols ``+,'' ``=,'' and ``-'' denote whether the HV results of the MaOEDA-IR with DR mechanism are statistically better than, equal to, or worse than that of the corresponding MaOEDA-IR without the diversity repairing mechanism with a significant level $5\%$, respectively.}
\label{hv_results_on_dtlz1_7_diversity}
\begin{center}
\begin{tabular}{c|c|c|c}
\hline
\multirow{2}{*}{Problem}&\multirow{2}{*}{$M$}&\multicolumn{2}{c}{Diversity Repairing}\\
\cline{3-4}
&&With&Without\\
\hline
\multirow{5}{*}{DTLZ1}
&3&\textbf{0.847(1.2E-3)}&0.843(1.2E-3)(=)\\
\cline{2-4}
&5&\textbf{0.975(4.5E-4)}&0.933(6.6E-2)(+)\\
\cline{2-4}
&8&\textbf{0.997(1.7E-4)}&0.844(5.6E-2)(+)\\
\cline{2-4}
&10&\textbf{0.997(1.5E-4)}&0.590(7.5E-2)(+)\\
\cline{2-4}
&15&\textbf{0.999(1.2E-4)}&0.727(9.7E-3)(+)\\
\hline
\multirow{5}{*}{DTLZ2}
&3&\textbf{0.567(2.8E-3)}&0.503(4.3E-2)(+)\\
\cline{2-4}
&5&\textbf{0.796(4.0E-3)}&0.709(5.0E-2)(+)\\
\cline{2-4}
&8&\textbf{0.923(9.6E-4)}&0.811(9.2E-3)(+)\\
\cline{2-4}
&10&\textbf{0.943(7.2E-4)}&0.924(1.5E-2)(+)\\
\cline{2-4}
&15&\textbf{0.987(1.0E-3)}&0.933(1.2E-3)(+)\\
\hline
\multirow{5}{*}{DTLZ3}
&3&\textbf{0.566(1.3E-3)}&0.558(1.0E-3)(=)\\
\cline{2-4}
&5&\textbf{0.999(4.1E-6)}&0.777(3.1E-2)(+)\\
\cline{2-4}
&8&\textbf{0.997(5.5E-6)}&0.785(2.0E-2)(+)\\
\cline{2-4}
&10&\textbf{0.998(2.8E-3)}&0.554(6.3E-2)(+)\\
\cline{2-4}
&15&\textbf{0.982(6.1E-4)}&0.652(1.7E-4)(+)\\
\hline
\multirow{5}{*}{DTLZ4}
&3&\textbf{0.750(8.3E-2)}&0.745(3.9E-3)(=)\\
\cline{2-4}
&5&\textbf{0.937(1.7E-2)}&0.925(3.4E-2)(+)\\
\cline{2-4}
&8&\textbf{0.991(2.9E-3)}&0.881(5.5E-2)(+)\\
\cline{2-4}
&10&\textbf{0.992(3.8E-3)}&0.813(4.9E-2)(+)\\
\cline{2-4}
&15&\textbf{0.999(6.1E-4)}&0.900(3.7E-5)(+)\\
\hline

\multirow{5}{*}{DTLZ1$^-$}
&3&\textbf{0.338(3.9E-3)}&0.285(3.1E-2)(+)\\
\cline{2-4}
&5&\textbf{0.011(3.3E-4)}&0.002(5.3E-2)(+)\\
\cline{2-4}
&8&\textbf{0.142(5.1E-3)}&0.129(3.0E-3)(+)\\
\cline{2-4}
&10&\textbf{0.124(7.0E-3)}&0.042(5.1E-2)(+)\\
\cline{2-4}
&15&\textbf{0.102(6.7E-3)}&0.010(2.8E-3)(+)\\
\hline
\multirow{5}{*}{DTLZ2$^-$}
&3&\textbf{0.541(1.4E-3)}&0.443(9.0E-2)(+)\\
\cline{2-4}
&5&\textbf{0.071(9.6E-4)}&0.063(3.0E-2)(+)\\
\cline{2-4}
&8&\textbf{0.103(1.5E-2)}&0.014(9.6E-2)(+)\\
\cline{2-4}
&10&\textbf{0.038(7.7E-3)}&0.006(2.9E-2)(+)\\
\cline{2-4}
&15&\textbf{0.003(8.9E-4)}&0.000(0.0E-0)(=)\\
\hline
\multirow{5}{*}{DTLZ3$^-$}
&3&\textbf{0.595(2.1E-2)}&0.518(9.1E-2)(+)\\
\cline{2-4}
&5&\textbf{0.255(3.0E-2)}&0.240(6.1E-2)(+)\\
\cline{2-4}
&8&\textbf{0.001(6.5E-5)}&0.000(0.0E-0)(=)\\
\cline{2-4}
&10&\textbf{0.036(7.2E-3)}&0.013(1.1E-4)(+)\\
\cline{2-4}
&15&\textbf{0.003(9.0E-4)}&0.000(0.0E-0)(=)\\
\hline
\multirow{5}{*}{DTLZ4$^-$}
&3&\textbf{0.540(1.2E-3)}&0.472(9.7E-4)(+)\\
\cline{2-4}
&5&\textbf{0.069(7.7E-4)}&0.031(9.7E-2)(+)\\
\cline{2-4}
&8&\textbf{0.001(5.9E-5)}&0.000(0.0E-0)(=)\\
\cline{2-4}
&10&\textbf{0.043(8.0E-3)}&0.010(5.5E-2)(+)\\
\cline{2-4}
&15&\textbf{0.005(1.5E-3)}&0.000(0.0E-0)(=)\\
\hline
\multicolumn{2}{c|}{+/=/-}&&33/7/0 \\
\hline
\end{tabular}
\end{center}
\end{table}

Furthermore, Table~\ref{hv_results_on_dtlz1_7_diversity} shows the extensive experimental comparisons between the proposed algorithm with and without the diversity repairing mechanism. Specifically, these experiments are independently performed $30$ runs over each test problem, and their results are measured by HV. Moreover, the best median HV results are highlighted in bold face, and the symbols ``+,'' ``=,'' and ``-'' denote whether the HV results of the proposed algorithm with the diversity repairing mechanism are statistically better than, equal to, or worse than those of the proposed algorithm without the diversity repairing mechanism with a significant level $5\%$, respectively. In addition, the last row in Table 3 summarizes how many times the proposed algorithm with the diversity repairing mechanism are better than, equal to, or worse than itself without this mechanism. It is clearly shown in Table~\ref{hv_results_on_dtlz1_7_diversity} that when the diversity repairing mechanism is employed, the proposed algorithm obtains all the best median HV results and most best statistical results against its competitors over DTLZ1-DTLZ4, and DTLZ1$^-$-DTLZ4$^-$ with $3$-, $5$-, $8$-, $10$-, and $15$-objective, while  the proposed algorithm without the diversity repairing mechanism could not even obtain effective solutions over DTLZ2$^-$ with $15$-objective, DTLZ3$^-$ with 8- and 15-objective, and DTLZ4$^-$ with 8-objective (i.e., the HV results upon these generated solutions are approximately zeros, which is caused by the domination by the employed reference points for the calculation of HV). In addition, it also can be observed that the diversity repairing mechanism may not significantly improve the performance in solving MOPs where the phenomenon of diversity losing is not severe. For example, the proposed algorithm obtains the same statistical results over 3-objective DTLZ1, DTLZ3, and DTLZ4 test problems no matter if the diversity repairing mechanism is employed. In summary, the diversity repairing mechanism can significantly improve the performance of the proposed algorithm especially in solving MaOPs.

\subsection{Investigation on Dimension Reduction}
\label{sec_investigation_dimension_reduction}
It is expected that dimension reduction in the decision variable space is capable of reducing the computational complexity of the proposed MaOEDA-IR. In this situation, two types of experiments are to be performed in order to draw any meaningful conclusion. The first one is to measure the performance of the solutions generated by the proposed algorithm with and without the dimension reduction within the same generation numbers. The other one is to compare the generation numbers when the same performance is achieved by the proposed algorithm with and without the dimension reduction. In the following, the first experimental results would be shown, while the second experimental comparisons are presented in Supplemental Materials. Specifically, the generation number of the first experiment is adopted from the parameter settings in Subsection~\ref{sec_parameter_settings} (i.e., $200$). The experimental comparisons are independently performed $30$ runs by the proposed algorithm with and without the dimension reduction over DTLZ1-DTLZ4 and DTLZ1$^-$-DTLZ4$^-$ with $3$-, $5$-, $8$-, $10$-, and $15$-objective. Then their results are measured by HV and shown in Table~\ref{hv_results_on_dtlz1_7_dimension} where the best median HV results are highlighted in bold face, and the symbols ``+,'' ``=,'' and ``-'' denote whether the HV results of the proposed algorithm with the dimension reduction are statistically better than, equal to, or worse than that of the proposed algorithm without the dimension reduction with a significant level $5\%$, respectively. Furthermore, the last row in Table~\ref{hv_results_on_dtlz1_7_dimension} summarizes how many times the proposed algorithm with the dimension reduction are better than, equal to, or worse than itself without this technique. It is obvious from Table~\ref{hv_results_on_dtlz1_7_dimension} that the proposed algorithm obtains the significant performance improvement when the dimension reduction is employed. Furthermore, without the dimension reduction, the proposed algorithm cannot perform well over several test problems, such as the DTLZ1$^-$ with $5$-objective, DTLZ2$^-$ with $10$- and $5$-objective, and DTLZ3$^-$ as well as DTLZ4$^-$ with $8$-, $10$-, and $15$-objective (their HV results are zeros). In summary, the proposed algorithm shows its superiority when the dimension reduction is utilized.

\begin{table}[htp]
\caption{HV results of MaOEDA-IR with and without the dimension reduction over the DTLZ1, DTLZ2, DTLZ3, DTLZ4, DTLZ1$^-$, DTLZ2$^-$, DTLZ3$^-$, and DTLZ4$^-$ test problems with $3$-, $5$-, $8$-, $10$-, and $15$-objective. Each compared algorithm is independently performed $30$ runs, and the best median HV results are highlighted in bold face. The symbols ``+,'' ``=,'' and ``-'' denote whether the HV results of the MaOEDA-IR with the dimension reduction are statistically better than, equal to, or worse than that of the corresponding MaOEDA-IR without the dimension reduction with a significant level $5\%$, respectively.}
\label{hv_results_on_dtlz1_7_dimension}
\begin{center}
\begin{tabular}{c|c|c|c}
\hline
\multirow{2}{*}{Problem}&\multirow{2}{*}{$M$}&\multicolumn{2}{c}{ Dimension Reduction}\\
\cline{3-4}
&&With&Without\\
\hline
\multirow{5}{*}{DTLZ1}
&3&\textbf{0.847(1.2E-3)}&0.776(3.2E-3)(+)\\
\cline{2-4}
&5&\textbf{0.975(4.5E-4)}&0.969(3.4E-4)(+)\\
\cline{2-4}
&8&\textbf{0.997(1.7E-4)}&0.902(4.4E-3)(+)\\
\cline{2-4}
&10&\textbf{0.997(1.5E-4)}&0.948(6.6E-2)(+)\\
\cline{2-4}
&15&\textbf{0.999(1.2E-4)}&0.954(1.6E-2)(+)\\
\hline
\multirow{5}{*}{DTLZ2}
&3&\textbf{0.567(2.8E-3)}&0.555(2.6E-2)(+)\\
\cline{2-4}
&5&\textbf{0.796(4.0E-3)}&0.746(5.1E-2)(+)\\
\cline{2-4}
&8&\textbf{0.923(9.6E-4)}&0.907(7.0E-3)(+)\\
\cline{2-4}
&10&\textbf{0.943(7.2E-4)}&0.909(8.9E-2)(+)\\
\cline{2-4}
&15&\textbf{0.987(1.0E-3)}&0.828(9.6E-3)(+)\\
\hline
\multirow{5}{*}{DTLZ3}
&3&\textbf{0.566(1.3E-3)}&0.511(2.4E-5)(+)\\
\cline{2-4}
&5&\textbf{0.999(4.1E-6)}&0.985(9.3E-3)(+)\\
\cline{2-4}
&8&\textbf{0.997(5.5E-6)}&0.912(3.5E-4)(+)\\
\cline{2-4}
&10&\textbf{0.998(2.8E-3)}&0.328(2.0E-2)(+)\\
\cline{2-4}
&15&\textbf{0.982(6.1E-4)}&0.928(2.5E-4)(+)\\
\hline
\multirow{5}{*}{DTLZ4}
&3&\textbf{0.750(8.3E-2)}&0.688(2.9E-4)(+)\\
\cline{2-4}
&5&\textbf{0.937(1.7E-2)}&0.890(7.6E-2)(+)\\
\cline{2-4}
&8&\textbf{0.991(2.9E-3)}&0.956(7.5E-3)(+)\\
\cline{2-4}
&10&\textbf{0.992(3.8E-3)}&0.109(3.8E-4)(+)\\
\cline{2-4}
&15&\textbf{0.999(6.1E-4)}&0.940(5.7E-4)(+)\\
\hline

\multirow{5}{*}{DTLZ1$^-$}
&3&\textbf{0.338(3.9E-3)}&0.330(4.7E-2)(+)\\
\cline{2-4}
&5&\textbf{0.011(3.3E-4)}&0.000(0.0E-0)(=)\\
\cline{2-4}
&8&\textbf{0.142(5.1E-3)}&0.089(3.4E-2)(+)\\
\cline{2-4}
&10&\textbf{0.124(7.0E-3)}&0.046(1.6E-4)(+)\\
\cline{2-4}
&15&\textbf{0.102(6.7E-3)}&0.009(7.9E-3)(+)\\
\hline
\multirow{5}{*}{DTLZ2$^-$}
&3&\textbf{0.541(1.4E-3)}&0.510(7.5E-4)(+)\\
\cline{2-4}
&5&\textbf{0.071(9.6E-4)}&0.018(4.5E-2)(+)\\
\cline{2-4}
&8&\textbf{0.103(1.5E-2)}&0.086(8.4E-3)(+)\\
\cline{2-4}
&10&\textbf{0.038(7.7E-3)}&0.000(0.0E-0)(=)\\
\cline{2-4}
&15&\textbf{0.003(8.9E-4)}&0.000(0.0E-0)(=)\\
\hline
\multirow{5}{*}{DTLZ3$^-$}
&3&\textbf{0.595(2.1E-2)}&0.580(9.6E-3)(+)\\
\cline{2-4}
&5&\textbf{0.255(3.0E-2)}&0.172(4.6E-5)(+)\\
\cline{2-4}
&8&\textbf{0.001(6.5E-5)}&0.000(0.0E-0)(=)\\
\cline{2-4}
&10&\textbf{0.036(7.2E-3)}&0.000(0.0E-0)(=)\\
\cline{2-4}
&15&\textbf{0.003(9.0E-4)}&0.000(0.0E-0)(=)\\
\hline
\multirow{5}{*}{DTLZ4$^-$}
&3&\textbf{0.540(1.2E-3)}&0.532(2.6E-3)(+)\\
\cline{2-4}
&5&\textbf{0.069(7.7E-4)}&0.029(1.5E-3)(+)\\
\cline{2-4}
&8&\textbf{0.001(5.9E-5)}&0.000(0.0E-0)(=)\\
\cline{2-4}
&10&\textbf{0.043(8.0E-3)}&0.000(0.0E-0)(=)\\
\cline{2-4}
&15&\textbf{0.005(1.5E-3)}&0.000(0.0E-0)(=)\\
\hline
\multicolumn{2}{c|}{+/=/-}&&31/9/0 \\
\hline
\end{tabular}
\end{center}
\end{table}

\section{Conclusion}
\label{section_5}
In solving many-objective optimization problems, the performance of most multi-objective evolutionary algorithms often deteriorate appreciably because of the loss of selection pressure during the evolution process. This is largely due to the selected parents solutions not generating promising individuals with the conventional genetic operators to direct the search towards the Pareto-optimal front. An improved regularity-based estimation of distribution algorithm, which generates new solutions with a probabilistic model built from the solutions the algorithm has visited, is proposed in this paper. To be specific, the proposed algorithm made an innovation in the following aspects: 1) devising a diversity repairing mechanism to reduce the risk of dominance resistant solutions and 2) generating promising solutions with the statistics of regularity learnt from the neighboring solutions with respect to the representatives which are uniformly distributed in the objective space. These two steps works in conjunction with each other to direct the search towards Pareto-optimal front. In addition, dimension reduction technique is utilized to reduce the cost of exploitation and exploration. Furthermore, in addition to the investigations are performed on the diversity repairing and dimension reduction, investigation is also performed based on the size of neighbors affecting the performance of the proposed algorithm to give the guideline for decision-marker. Extensive experiments are performed and the results measured by the chosen performance metrics indicate that the proposed algorithm shows superiority in tackling many-objective optimization problems. In our future research, we will extend the proposed algorithm to deal with highly constrained many-objective optimization problems in which complicated regularity of Pareto fronts often exists.

\IEEEpeerreviewmaketitle
\ifCLASSOPTIONcaptionsoff
  \newpage
\fi
\bibliographystyle{IEEEtran}

\begin{IEEEbiography}[{\includegraphics[width=1in,height=1.25in,clip,keepaspectratio]{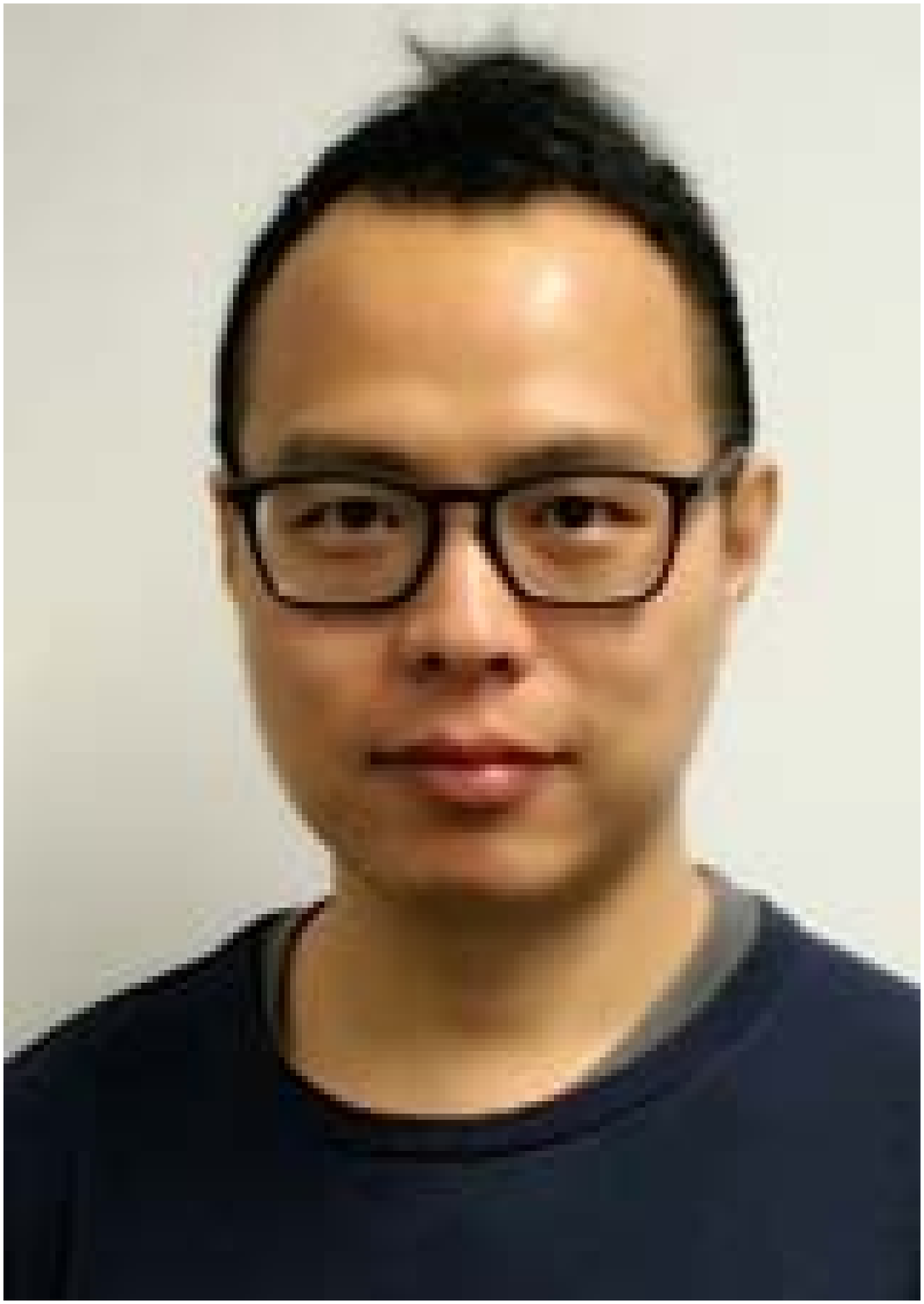}}]{Yanan Sun} (S'15-M'18) received a Ph.D. degree in engineering from the Sichuan University, Chengdu, China, in 2017. From 2015.08-2017.02, he is a jointly Ph.D. student financed by the China Scholarship Council in the School of Electrical and Computer Engineering, Oklahoma State University (OSU), USA. He is currently a Postdoc Research Fellow in the School of Engineering and Computer Science, Victoria University of Wellington, Wellington, New Zealand. His research topics are many-objective optimization and deep learning.
\end{IEEEbiography}

\begin{IEEEbiography}[{\includegraphics[width=1in,height=1.25in,clip,keepaspectratio]{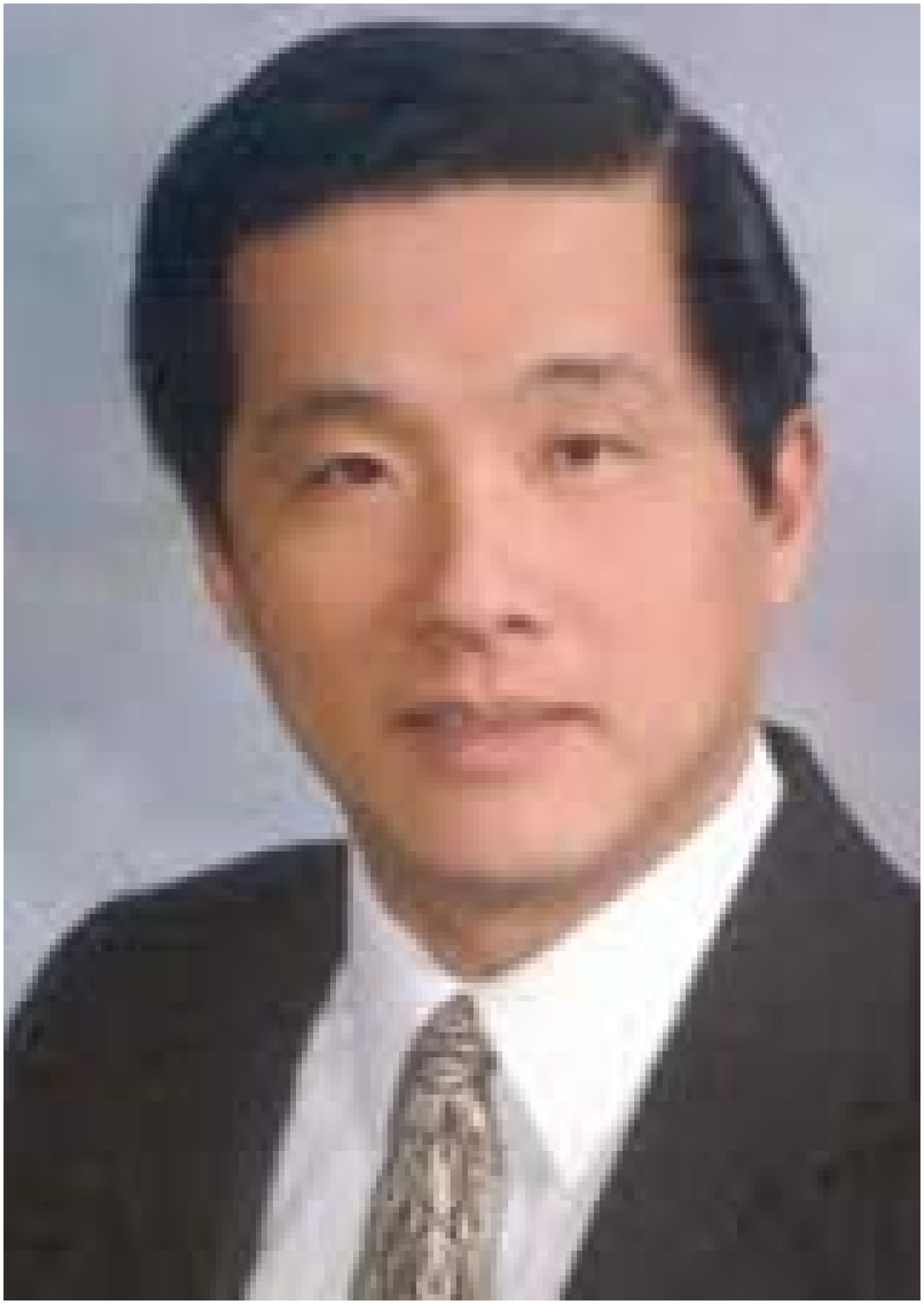}}]{Gary G. Yen}
	(S'87-M'88-SM'97-F'09) received a Ph.D. degree in electrical and computer engineering from the University of Notre Dame in 1992. Currently he is a Regents Professor in the School of Electrical and Computer Engineering, Oklahoma State University (OSU). Before joined OSU in 1997, he was with the Structure Control Division, U.S. Air Force Research Laboratory in Albuquerque. His research interest includes intelligent control, computational intelligence, conditional health monitoring, signal processing and their industrial/defense applications.
	
	Dr. Yen was an associate editor of the \textit{IEEE Control Systems Magazine, IEEE Transactions on Control Systems Technology}, \textit{Automatica}, \textit{Mechantronics}, \textit{IEEE Transactions on Systems, Man and Cybernetics, Parts A and B} and I\textit{EEE Transactions on Neural Networks}. He is currently serving as an associate editor for the \textit{IEEE Transactions on Evolutionary Computation} and the \textit{IEEE Transactions on Cybernetics}. He served as the General Chair for the \textit{2003 IEEE International Symposium on Intelligent Control} held in Houston, TX and \textit{2006 IEEE World Congress on Computational Intelligence} held in Vancouver, Canada. Dr. Yen served as Vice President for the Technical Activities in 2005-2006 and then President in 2010-2011 of the IEEE Computational intelligence Society. He was the founding editor-in-chief of the \textit{IEEE Computational Intelligence Magazine}, 2006-2009. In 2011, he received Andrew P Sage Best Transactions Paper award from \textit{IEEE Systems, Man and Cybernetics Society} and in 2014, he received Meritorious Service award from \textit{IEEE Computational Intelligence Society}.
\end{IEEEbiography}

\begin{IEEEbiography}[{\includegraphics[width=1in,height=1.25in,clip,keepaspectratio]{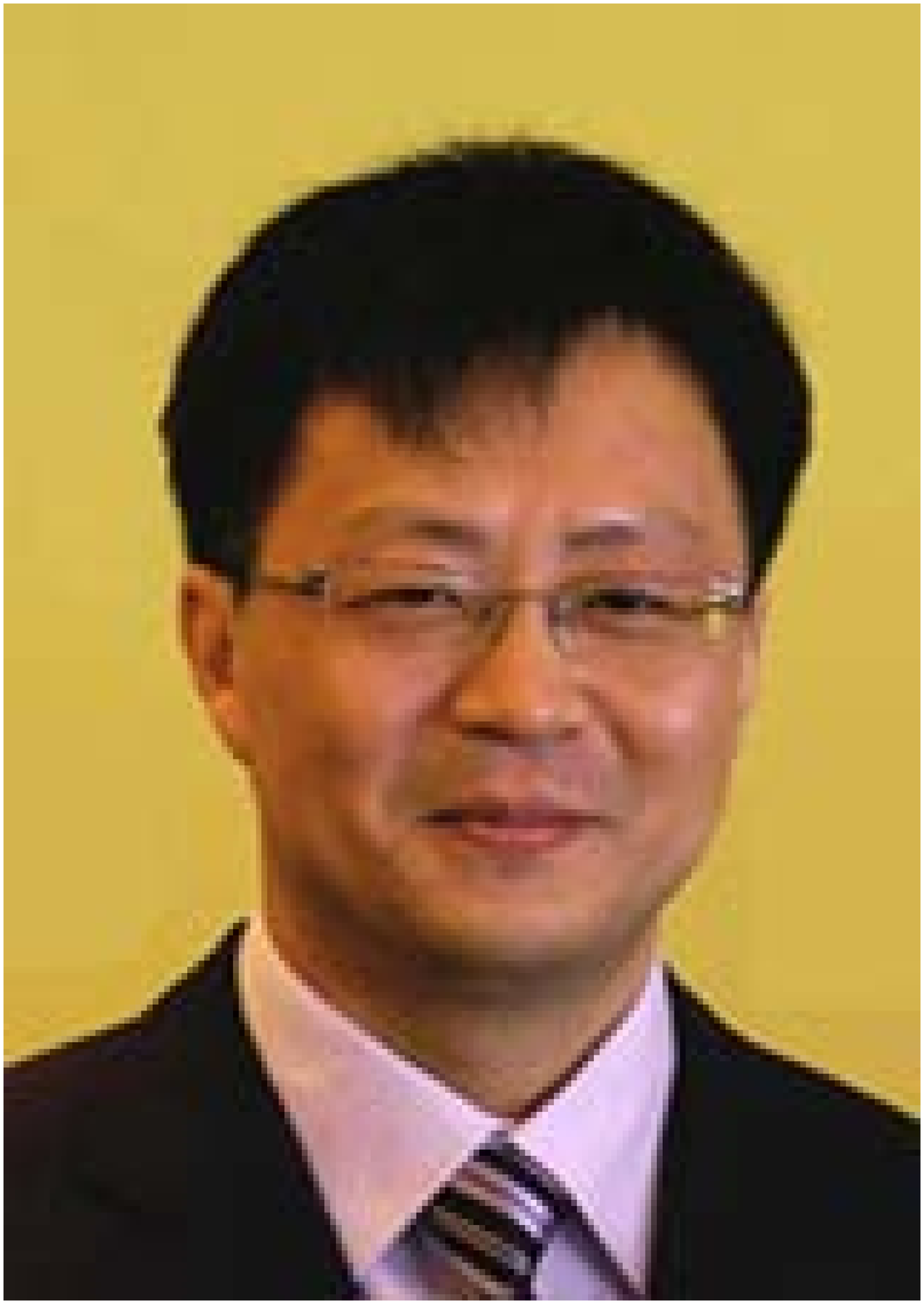}}]{Zhang Yi} (F'16)
	received a Ph.D. degree in mathematics from the Institute of Mathematics, The Chinese Academy of Science, Beijing, China, in 1994. Currently, he is a Professor at the Machine Intelligence Laboratory, College of Computer Science, Sichuan University, Chengdu, China. He is the co-author of three books: \emph{Convergence Analysis of Recurrent Neural Networks} (Kluwer Academic Publishers, 2004), \emph{Neural Networks: Computational Models and Applications} (Springer, 2007), and \emph{Subspace Learning of Neural Networks} (CRC Press, 2010). He was an Associate Editor of \emph{IEEE Transactions on Neural Networks and Learning Systems} (2009~2012), and He is an Associate Editor of \emph{IEEE Transactions on Cybernetics} (2014~). His current research interests include Neural Networks and Big Data. He is a fellow of IEEE.
\end{IEEEbiography}

\appendices
\section{}
\label{appendex_1}

In this section, experiments are designed to verify the performance of the dimension reduction in the proposed algorithm by counting the generation numbers when the same performance appears for the first time. Specifically, we first record the generation number when the baseline results\footnote{We first employ PCSEA to obtain the training data by setting its population size to be $100$ for $M\leq 10$ and $200$ for $M >10$, and then perform MaOEDA-IR to get the final solutions. The maximal generation numbers for PCSEA and MaOEDA-IR are set to be $200$. The HV of the final solutions are considered as the baseline results.} are reached for the first time. Then re-perform the proposed algorithm without the dimension reduction until the resulting $E$ satisfy the criterion formulated by Equation~(\ref{equ_stop_condition_exp}) or the generation number is greater than $500$.
\begin{equation}
\label{equ_stop_condition_exp}
\sqrt{\frac{(E-R)^2}{R^2}} \leq 0.0001
\end{equation}
where $R$ denotes the corresponding baseline results. 

\begin{figure}[htp]
	\centering
	\includegraphics[width=\columnwidth]{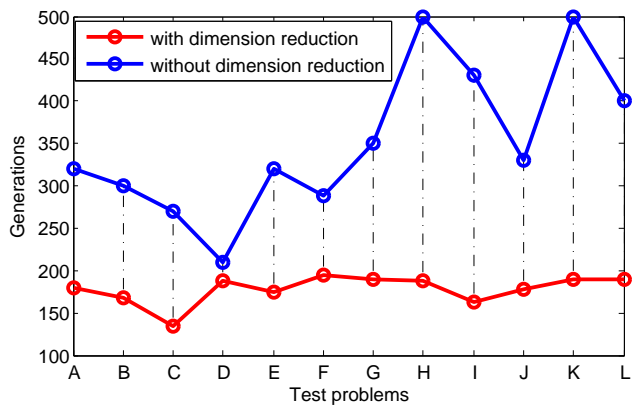}\\
	\caption{The comparisons between the proposed MaOEDA-IR with and without dimension reduction on DTLZ1-DTLZ4 with $8$-, $10$-, and $15$-objective test problems. Specifically, A, B, C in the $x$ axis denote $8$-, $10$-, and $15$-objective DTLZ1 problem, while D, E, F for DTLZ2, G, H, I for DTLZ3 and J, K, L for DTLZ4 test problems.}\label{fig_with_and_without_dr}
\end{figure}

Fig.~\ref{fig_with_and_without_dr} illustrates the comparisons over the $8$-, $10$-, and $15$-objective DTLZ1-DTLZ4 test problems which are denoted by A-L, and the red line and blue line denote the results generated by the proposed MaOEDA-IR with and without dimension reduction, respectively. The results in Fig.~\ref{fig_with_and_without_dr} clearly shows the generation numbers performed by the MaOEDA-IR without dimension reduction are about twice as many to that of MaOEDA-IR with dimension reduction except on the $8$-objective DTLZ2 test problem (denoted by ``D''). In addition, MaOEDA-IR without dimension reduction cannot obtain the results satisfying Equation~(\ref{equ_stop_condition_exp}) over $10$-objective  DTLZ3 and DTLZ4 test problems until the maximum pre-defined generation number is reached (denoted by ``H'' and ``K'', respectively). In summary, experiments in Fig.~\ref{fig_with_and_without_dr} validate our motivation of employing the dimension reduction in the proposed MaOEDA-IR.
\end{document}